\documentclass{article}
\usepackage{ijcai23}

\usepackage{times}
\usepackage{soul}
\usepackage{url}
\usepackage[hidelinks]{hyperref}
\usepackage[utf8]{inputenc}
\usepackage[small]{caption}
\usepackage{graphicx}
\usepackage{amsmath}
\usepackage{amsthm}
\usepackage{booktabs}
\usepackage{algorithm}
\usepackage{algorithmic}
\usepackage[switch]{lineno}

\usepackage{mathtools}
\usepackage{subcaption}
\usepackage{booktabs}
\usepackage{verbatim}
\usepackage{fancyvrb}
\usepackage{xcolor}

\title{Optimal Decision Tree Policies for Markov Decision Processes}

\author{
Dani\"el Vos
\and
Sicco Verwer
\affiliations
Delft University of Technology
\emails
\{d.a.vos, s.e.verwer\}@tudelft.nl
}

\begin{document}

\maketitle

\begin{abstract}
Interpretability of reinforcement learning policies is essential for many real-world tasks but learning such interpretable policies is a hard problem. Particularly, rule-based policies such as decision trees and rules lists are difficult to optimize due to their non-differentiability. While existing techniques can learn verifiable decision tree policies, there is no guarantee that the learners generate a policy that performs optimally. In this work, we study the optimization of size-limited decision trees for Markov Decision Processes (MPDs) and propose OMDTs: Optimal MDP Decision Trees. Given a user-defined size limit and MDP formulation, OMDT directly maximizes the expected discounted return for the decision tree using Mixed-Integer Linear Programming. By training optimal tree policies for different MDPs we empirically study the optimality gap for existing imitation learning techniques and find that they perform sub-optimally. We show that this is due to an inherent shortcoming of imitation learning, namely that complex policies cannot be represented using size-limited trees. In such cases, it is better to directly optimize the tree for expected return. While there is generally a trade-off between the performance and interpretability of machine learning models, we find that on small MDPs, depth 3 OMDTs often perform close to optimally.
\end{abstract}

\section{Introduction}
Advances in reinforcement learning using function approximation have allowed us to train powerful agents for complex problems such as the games of Go and Atari~\cite{schrittwieser2020mastering}.
Policies learned using function approximation often use neural networks, making them impossible for humans to understand.
Therefore reinforcement learning is severely limited for applications with high-stakes decisions where the user has to trust the learned policy.

Recent work has focused on \textit{explaining} opaque models such as neural networks by attributing prediction importance to the input features~\cite{ribeiro2016should,lundberg2017unified}. However, these explanation methods cannot capture the full complexity of their models, which can mislead users when attempting to understand the predictions \cite{rudin2019stop}. Concurrently, there has been much work on \textit{interpretable} machine learning in which the model learned is limited in complexity to the extent that humans can understand the complete model. Particularly decision trees have received much attention as they are simple models that are capable of modeling non-linear behavior~\cite{lipton2018mythos}. 

Decision trees are difficult to optimize as they are non-differentiable and discontinuous. Previous works have used different strategies to overcome the hardness of optimizing trees: using assumptions or relaxations to make the trees differentiable~\cite{gupta2015policy,silva2020optimization,likmeta2020combining}, reformulating the MDP into a meta-MDP that exclusively models decision tree policies~\cite{topin2021iterative} or extracting trees from a complex teacher~\cite{bastani2018verifiable}. While these methods are increasingly successful in training performant trees they do not offer guarantees on this performance.

\begin{figure}[b]
    \centering
    \includegraphics[width=\linewidth]{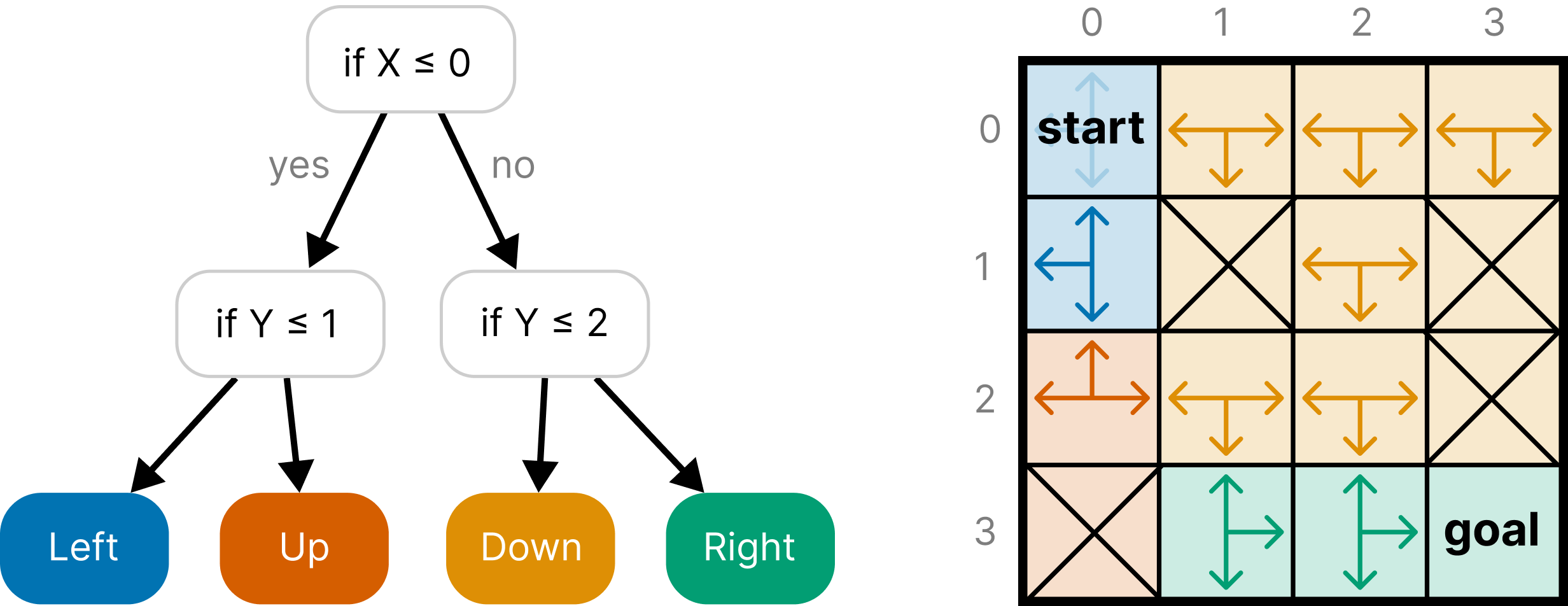}
    \caption{Depth 2 OMDT on the stochastic Frozenlake 4x4 environment. OMDT proves that no better depth 2 decision tree policy exists (discounted return $0.37$ with $\gamma=0.99$).}
    \label{fig:omdt-example}
\end{figure}

Our work takes a first step at bridging the gap between the fields of optimal decision trees and reinforcement learning. Existing formulations for optimal decision trees assume a fixed training set with independent samples. This cannot be used in a dynamic setting where actions taken in one state influence the best actions in others. Instead, we formulate the problem of solving a Markov Decision Process (MDP) using a policy represented by a size-limited decision tree (see Figure~\ref{fig:omdt-example}) in a single MILP. We link the predictions of the decision tree policy to the state-action frequencies in the dual linear program for solving MDPs. The dual allows us to reason over policies explicitly, which results in a more efficient formulation.
Our formulation for Optimal MDP Decision Trees, OMDTs, optimizes a decision tree policy for a given MDP and a tree size limit. OMDT produces increasingly performant policies as runtime progresses and eventually proves the optimality of its policy under the size constraint.

Existing methods for training size-limited decision trees in reinforcement learning such as VIPER~\cite{bastani2018verifiable} make use of imitation learning, where a student tries to learn from a powerful teacher policy. We compare the performance of OMDT and VIPER on a variety of MDPs. Interestingly, we show that, when training interpretable size-limited trees, imitation learning performs significantly worse as capacity of the learned decision tree is wasted on parts of the state space that are never reached by the policy. Moreover, VIPER cannot prove optimality even if it identifies the optimal solution. Regarding the performance-interpretability trade-off, we show that decision trees of 7 decision nodes are enough to perform close to unrestricted optimal policies in 8 out of 13 environments. Such trees are orders of magnitude smaller than size-unrestricted trees created by methods that replicate the unrestricted policy such as dtcontrol~\cite{ashok2020dtcontrol}.

\section{Background}

\subsection{Decision Trees}
Decision trees~\cite{breiman1984classification,quinlan1986induction} are simple models that execute a series of comparisons between a feature value and a threshold in the nodes to arrive at a leaf node that contains the prediction value. Due to their simple descriptions, size-limited decision trees are easy to understand for humans~\cite{molnar2020interpretable}. Particularly, size-limited decision trees admit \textit{simulatability}~\cite{lipton2018mythos}: humans can use the model to make predictions by hand in reasonable time and \textit{decomposability}: humans can understand each individual aspect of the model. The method proposed in this paper, OMDT, also offers \textit{algorithmic transparency}: we can trust the learning algorithm to produce models that fulfill certain qualities such as global optimality. Therefore decision trees are an attractive model class when interpretable policies are required.

\subsection{Markov Decision Processes}
Markov Decision Processes (MDPs)~\cite{bellman1957markovian} are the processes underlying reinforcement learning problems. An MDP models a decision-making process in a stochastic environment where the agent has some control over the state transitions. An MDP can be described by a tuple $\langle S, A, P, R \rangle$, where $S$ contains all states, $A$ the set of actions an agent can take, $P_{s,s',a}$ the probabilities of transitioning from state $s$ to state $s'$ when taking action $a$, and $R_{s,s',a}$ the reward the agent receives when going from state $s$ to state $s'$ under action $a$. When solving an MDP, we want to find a policy $\pi: S \rightarrow A$ such that, when executing its actions, the expected sum of rewards (the return) is maximized. In this work we define policies (w.l.o.g.) as a mapping from states and actions to an indicator for whether or not action $a$ is taken in state $s$: $\pi: S \times A \rightarrow \{0, 1\}$.

\subsubsection{Value Iteration} When solving MDPs we generally discount future rewards in each step by a user-defined value of $0 < \gamma < 1$ to ensure that the optimal policy will generate a finite return. The most common approach for optimally solving MDPs is by using one of many dynamic programming variants. In this work, we focus on value iteration. Value iteration finds a value $V_s$ for each state $s$ that holds the expected discounted return for taking optimal greedy actions starting from that state. These values can be found by iteratively updating $V_s$ until the Bellman equation~\cite{bellman1957markovian}
\begin{equation*}
    V_s = \sum_{s'} P_{s, s', a} R_{s, s', a} + \sum_{s'} \gamma P_{s, s', a} V_{s'} 
\end{equation*}
is approximately satisfied. We will refer to this optimal solution found with value iteration as the unrestricted optimal solution as the computed policy can be arbitrarily complex.

\section{Related Work}

\subsection{Learning Decision Tree Policies}
Decision trees have appeared at various parts of the reinforcement learning pipeline~\cite{glanois2021survey} for example in modeling the Q-value function or in modeling the policy. In this work, we are interested in modeling the policy with a decision tree as this gives us an interpretable model that we can directly use at test time.

Decision trees make predictions using hard comparisons between a single feature value and a threshold. The learned models can be discontinuous and non-differentiable, which makes optimization with gradients challenging. One line of research focuses on overcoming the non-differentiability of decision trees to allow for the use of gradient-based optimization methods. Gupta et al.~[\citeyear{gupta2015policy}] train decision trees that contain linear models in their leaves that can be optimized with gradients but this results in models that are hard to interpret.
Silva et al.~[\citeyear{silva2020optimization}] first relax the constraints that decision nodes select one feature, that leaves predict one value, and that thresholds are hard step functions. Such relaxed trees are differentiable and can be trained with policy gradient methods. By discretizing the relaxed tree they end up with an interpretable model that approximates the relaxed model. However, the relaxed tree can get stuck in local minima and the discretized tree offers no performance guarantees.
Likmeta et al.~[\citeyear{likmeta2020combining}] consider decision tree policies for autonomous driving tasks. To make the decision tree parameters easily optimizable, they fix the tree structure along with the features used in the decision nodes.
By learning a differentiable hyper-policy over decision tree policies they are then able to approximately optimize the models with gradient descent.

With Iterative Bounding MDPs, Topin et al.~[\citeyear{topin2021iterative}] reformulate the underlying MDP into one where the agent implicitly learns a decision tree policy. The method can be thought of as a tree agent learning to take actions that gather information and a leaf agent learning to take actions that work well given the gathered information of the tree agent. By reformulating the MDP, its decision tree policy can be optimized using differentiable function approximators and gradient-based optimizers.

In a separate line of work, the goal is to represent a specific, usually optimal, policy as a decision tree that is unbounded in size. These techniques have been developed for policies with a single goal state~\cite{brazdil2015counterexample} and as a tool for general controllers~\cite{ashok2020dtcontrol}: dtcontrol.

\subsubsection{Imitation Learning (VIPER)}
Instead of directly optimizing a decision tree, one can also try to extract a decision tree policy from a more complex teacher policy using imitation learning. These imitation learning algorithms turn reinforcement learning into a supervised learning problem for which we have successful decision tree learning algorithms~\cite{breiman1984classification,quinlan1986induction}. DAGGER~\cite{ross2011reduction} (dataset aggregation) is an algorithm that iteratively collects traces from the environment using its current policy and trains a supervised model on the union of the current and previous traces. Since DAGGER only uses information on the predicted action of the teacher policy, it ignores extra information on Q-values that modern Q-learning algorithms provide. VIPER~\cite{bastani2018verifiable} focuses on learning decision trees and improves on DAGGER by including Q-value information into the supervised learning objective. While VIPER generates significantly smaller decision trees than DAGGER, we will show that these trees are not yet optimal with respect to the trade-off in size and performance.

\subsection{Optimal Decision Trees}
The standard algorithms for training decision trees in supervised learning are greedy heuristics and can learn trees that perform arbitrarily poorly~\cite{kearns1996boosting}. Therefore in recent years there has been increasing interest in the design of algorithms that train decision trees to perform optimally. Early works formulated training decision trees for classification and regression and used methods such as dynamic programming to find optimal decision trees~\cite{nijssen2007mining}. Mixed-Integer Linear Programming based formulations~\cite{bertsimas2017optimal,verwer2017learning} have since become popular. These methods are flexible and have been extended to optimize performance under fairness constraints~\cite{aghaei2019learning} or directly optimize adversarial robustness~\cite{vos2022robust}. Generally, the size of the tree is limited to provide regularization and aid interpretability, then the solver is tasked with finding a decision tree that maximizes training performance given the size limits. The field has since worked on increasingly efficient optimization techniques using a variety of methods such as MILP~\cite{verwer2019learning}, dynamic programming~\cite{demirovic2020murtree,lin2020generalized}, constraint programming~\cite{verhaeghe2020learning}, branch-and-bound search~\cite{aglin2020learning,aglin2021pydl8} and Boolean (maximum) satisfiability~\cite{narodytska2018learning,hu2020learning,schidler2021sat}.

\section{OMDT: Optimal MDP Decision Trees}
As a first step in bridging the gap between optimal decision trees for supervised and reinforcement learning, we introduce OMDTs: Optimal MDP Decision Trees. 
OMDT is a Mixed-Integer Linear Programming formulation that encodes the problem of identifying a decision tree policy that achieves maximum discounted return given a user-defined MDP and tree size limit.
Our formulation can be solved using one of many available solvers, in this work we use the state-of-the-art solver Gurobi\footnote{\url{https://www.gurobi.com/}}.

\begin{figure}[t]
    \centering
    \includegraphics[width=\linewidth]{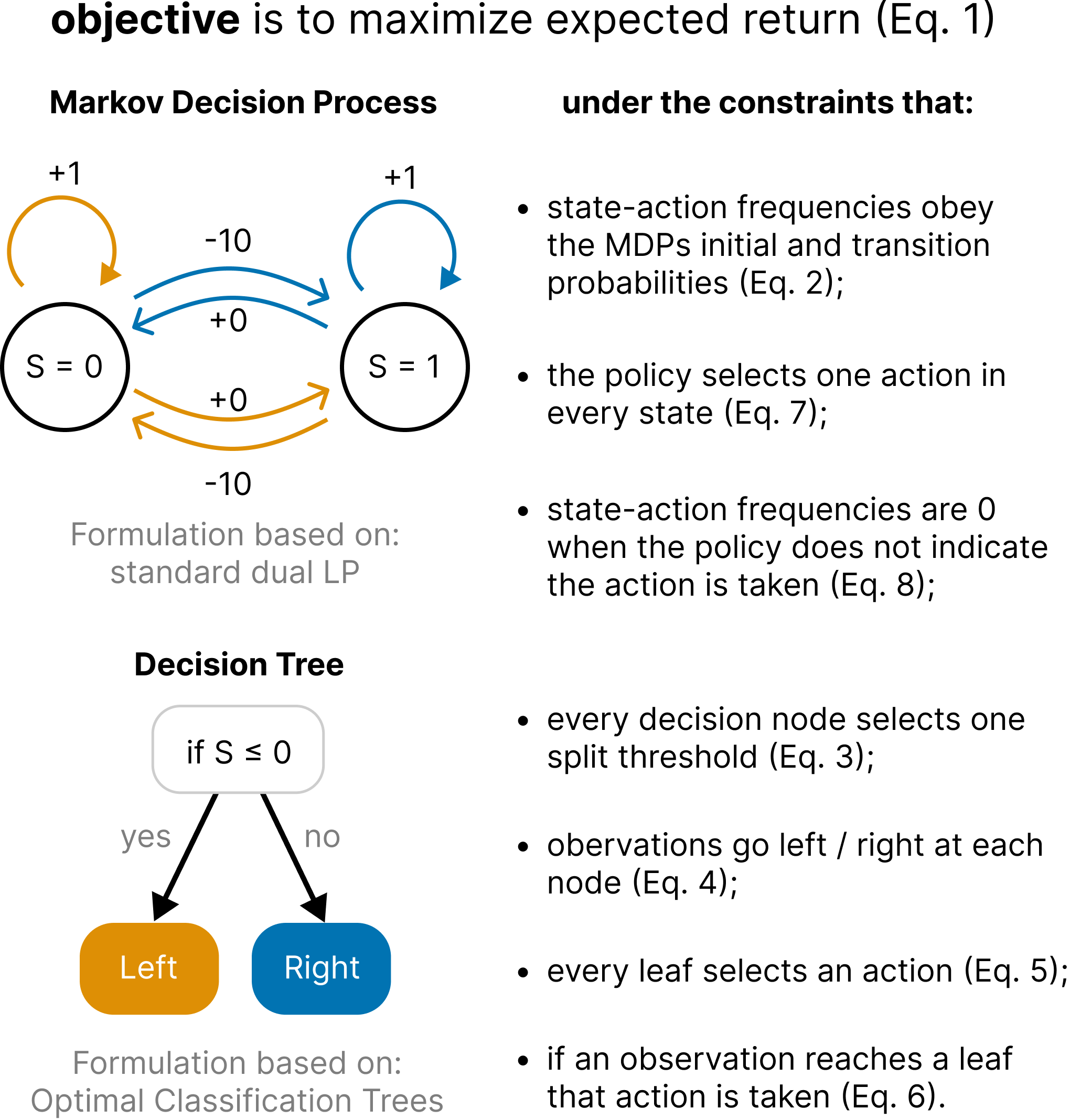}
    \caption{Overview of OMDT's formulation. We maximize the discounted return in an MDP under the constraint that the policy is represented by a size-limited decision tree.}
    \label{fig:omdt-intuition}
\end{figure}

Intuitively the OMDT formulation consists of two parts: a (dual) linear programming formulation for solving MDPs and a set of constraints that limits the set of feasible policies to decision trees. Figure \ref{fig:omdt-intuition} summarizes OMDT's formulation in natural language. All the notation used in OMDT is summarized in Table \ref{tab:notation}.

\subsection{Constraints}
It is well known that MDPs can be solved using linear programming, the standard linear program is~\cite{puterman2014markov}:
\begin{align}
    \text{min.}\;\; & \sum_s p_0(s) V_s \nonumber \\
    \text{s.t.}\;\; & V_s - \sum_{s'} \gamma P_{s, s', a} V_{s'} \geq \sum_{s'} P_{s, s', a} R_{s, s', a}, \;\; \forall s, a \nonumber 
\end{align}

It is not easy to efficiently add constraints to this formulation to enforce the policies to be size-limited decision trees because it reasons abstractly over policies, i.e. by reasoning over the policy's state values. To create a formulation for decision tree policies, we resort to the standard dual program:
\begin{align}
    \text{max.}\;\; & \sum_s \sum_a x_{s, a} \sum_{s'} P_{s, s', a} R_{s, s', a} \label{eq:dual-objective} \\
    \text{s.t.}\;\; & \sum_a x_{s, a} {-} \sum_{s'} \sum_a \gamma P_{s', s, a} x_{s', a} = p_0(s), \;\; \forall s \label{eq:dual-constraint}
\end{align}

This program uses a measure $x_{s, a}$ of how often the agent takes action $a$ in state $s$. This allows us to add efficient constraints that control the policy of the agent. Intuitively the program maximizes the rewards $\sum_{s'} P_{s,s',a} R_{s,s',a}$ weighted by this $x_{s,a}$. The constraints enforce that the frequency by which a state is exited is equal to the frequency that the agent is initialized in the state $p_0(s)$ or returns to it following the discounted transition probabilities $\gamma P_{s,s',a}$.

To enforce the policy to be a size-limited decision tree we will later constrain the $x_{s,a}$ values to only be non-zero when the agent is supposed to take action $a$ in state $s$ according to a tree policy. We will first introduce the variables and constraints required to model the decision tree constraints.

\subsubsection{Modeling Decision Nodes}
Our decision tree formulation is roughly based on OCT~\cite{bertsimas2017optimal} and ROCT~\cite{vos2022robust}, MILP formulations for optimal (OCT) and robust (ROCT) classification trees. In these formulations, the shape of the decision tree is fixed. 
Like ROCT, we describe a decision node $m$ by binary threshold variables $b_{m,j,k}$, indicating whether the $k$th threshold of feature $j$ is chosen.\footnote{In practice, the possible values for threshold $k$ depends on the chosen feature $j$. We do not model this for convenience of notation.} Unlike ROCT, we only allow one of these variables to be true over all features and possible thresholds:
\begin{equation}\label{eq:constraint-one-threshold}
    \sum_j \sum_k b_{m, j, k} = 1, \quad \forall m
\end{equation}

We follow paths through the tree to map observations to leaves. In each node $m$ we decide the direction $d_{s,m}$ that the observation of state $s$ takes (left=0 or right=1 of the threshold $k$). ROCT uses two variables per state-node pair to model directions $d_{s,m}$ to account for perturbations in the observations. Since we are optimizing for an MDP without uncertainty in the observations we only require one variable $d_{s,m}$ per state-node pair.

We further improve over ROCT by determining $d_{s,m}$ using only one constraint per state-node pair instead of a separate constraint per state-node-feature triple. For this, we pre-compute a function $\text{side}(s, j, k)$ which indicates for each feature-threshold pair $(j, k)$ and every observation $s$ the side of $k$ that $s$ is on for feature $j$ (left=0 or right=1), i.e. whether $X_{sj} > k$ holds. This formulation is not limited to the predicates `$\leq$' or `$>$' however and can be easily extended to other predicates in the pre-computation of $\text{side}(s, j, k)$. The following then forces the direction $d_{s,m}$ to be equal to the direction of the indicated threshold:

\begin{equation}
    d_{s, m} = \sum_j \sum_k \text{side}(s, j, k) \; b_{m, j, k}, \quad \forall s, m \label{eq:constraint-path}
\end{equation}

The variables $d_{s, m}$ represent the direction of an observation's path at a decision node. Together, the $d$ variables allow us to follow an observation's path through the tree which we use to identify the leaf that it reaches. Important in this formulation, compared to existing binary encodings, is that it requires no big-M constraints to describe these paths. This makes the relaxation stronger and therefore the solver gives much better bounds than using the big-M style formulations from ROCT.

\subsubsection{Modeling Policy Actions}
Decision leaves only have one set of binary decision variables: $c_{t, a}$ encoding whether or not leaf $t$ predicts action $a$. We want each leaf to select exactly one action:
\begin{equation} \label{eq:constraint-one-leaf-prediction}
    \sum_a c_{t, a} = 1, \quad \forall t \\
\end{equation}
As mentioned before we can follow an observation's path through the tree by using their $d_{s, m}$ path variables.
One can linearize an implication of a conjunction of binary variables as follows:
\begin{align*}
    x_1 \land x_2 \land ... \land x_n \implies y \quad \\
    \equiv x_1 + x_2 + ... + x_n - n + 1 \leq y 
\end{align*}
If an observation reaches leaf $t$ and the leaf predicts action $a$ then we want to force the policy $\pi_{s, a}$ to take that action in the associated state $s$. Using the aforementioned equivalence we add the constraint:
\begin{multline}
 \label{eq:constraint-policy-follows-tree}
    \;\;\; \sum_{\mathclap{m \in A_l(t)}} (1{-}d_{s, m}) + \sum_{\mathclap{m \in A_r(t)}} d_{s, m} \\ + c_{t, a} - |A(t)| \leq \pi_{s, a}, \quad
    \forall s, a, t
\end{multline}
This constraint forces the agent to take the action indicated by the leaf. To prevent the agent from taking other actions that were not indicated we force it to only take a single action in each state (giving a deterministic policy):
\begin{equation} \label{eq:constraint-deterministic-policy}
    \sum_a \pi_{s, a} = 1, \quad \forall s \\
\end{equation}
Now we have indicators $\pi_{s, a}$ that mark what action is taken by the agent. To link this back to the MDP linear programming formulation that we use to optimize the policy, we set the $x_{s, a}$ variables. We need to set $x_{s, a} = 0$ if $\pi_{s, a} = 0$, else $x_{s, a}$ should be unbounded. We encode this using a big-M formulation:
\begin{equation} \label{eq:constraint-frequency-follows-policy}
    x_{s, a} \leq M \pi_{s, a}, \quad \forall s, a \\
\end{equation}
$M$ should be chosen as small as possible, but larger or equal to the largest value that $x_{s, a}$ can take. We use the fact that we are optimizing the MDP using discount factor $\gamma$ to compute an upper bound on $x_{s, a}$ and set $M = \frac{1}{1 - \gamma}$, proof is given in the appendix.

\begin{table}[tb]
    \centering
    \setlength{\tabcolsep}{1.5pt}
    \begin{tabular}{@{}lll@{}}
        \toprule
        Name & Kind & Description \\
        \midrule
        $b_{m, j, k}$ & bin. & Tree uses feat. $j$ and threshold $k$ in node $m$ \\
        $c_{t, a}$ & bin. & Tree selects action $a$ in leaf $t$ \\
        $d_{s, m}$ & bin. & Observation of $s$ goes left / right in node $m$ \\
        $\pi_{s, a}$ & bin. & Policy takes action $a$ in state $s$ \\
        \midrule
        $x_{s, a}$ & cont. & Frequency of action $a$ taken in state $s$ \\
        \midrule
        $P_{s, s', a}$ & const. & Probability of transition $s{\rightarrow}s'$ with action $a$ \\
        $R_{s, s', a}$ & const. & Reward for transition $s{\rightarrow}s'$ with action $a$ \\
        $p_0(s)$ & const. & Probability of starting in state $s$ \\
        $\gamma$ & const. & Discount factor \\
        $X_{ij}$ & const. & Feature $j$'s value of observation $i$ \\
        \text{side}(s{,}j{,}k) & const. & Side state $s$ is on for thresh. $k$ and feat. $j$\\
        \midrule
        $a \in A$ & set & Set of actions in MDP \\
        $s \in S$ & set & Set of states in MDP \\
        $i {=} 1..|S|$ & set & Observation and state indices \\
        $j \in J$ & set & Set of feature indices \\
        $k=1..K$ & set & Indices of all possible feature thresholds\\
        $m \in \mathcal{T}_D$ & set & Set of decision nodes in the tree \\
        $t \in \mathcal{T}_L$ & set & Set of leaves in the tree \\
        $A(t)$ & set & Set of ancestors of leaf $t$ \\
        $A_{l}(t)$ & set & ... that have $t$ in their left path \\
        $A_{r}(t)$ & set & ... that have $t$ in their right path \\
        \bottomrule
    \end{tabular}
    \caption{Summary of notation used in OMDT.}
    \label{tab:notation}
\end{table}

\subsection{Complete Formulation}
The runtime of MILP solvers grows worst-case exponentially with respect to formulation size so it is important to limit the scale of the formulation. The number of variables in our formulation grows with $\mathcal{O}(|S||J||\mathcal{T}_D| {+} |A||\mathcal{T}_L|{+}|S||A|)$ which follows from their indices in Table \ref{tab:notation}. The number of constraints grows with the order $\mathcal{O}(|S||\mathcal{T}_D| {+} |S||A||\mathcal{T}_L|)$ as it is dominated by the constraints that determine $d_{s, m}$ at each node (Equation \ref{eq:constraint-path}) and constraints that force $\pi_{s,a}$ according to the tree (Equation \ref{eq:constraint-policy-follows-tree}). We summarize OMDT below:

\begin{equation}
    \max \quad \sum_s \sum_a x_{s, a} \sum_{s'} P_{s, s', a} R_{s, s', a} \tag{\ref{eq:dual-objective}}
\end{equation}
s.t.
\begin{align}
    \sum_a x_{s, a} - \sum_{s'} \sum_a \gamma P_{s', s, a} x_{s', a} = p_0(s),& \; \forall s \tag{\ref{eq:dual-constraint}} \\
    \sum_j \sum_k b_{m, j, k} = 1,& \; \forall m \tag{\ref{eq:constraint-one-threshold}} \\
    d_{s, m} = \sum_j \sum_k \text{side}(s, j, k) \; b_{m, j, k},& \; \forall s, m \tag{\ref{eq:constraint-path}} \\
    \sum_a c_{t, a} = 1,& \; \forall t \tag{\ref{eq:constraint-one-leaf-prediction}} \\
    \sum_{\mathclap{m \in A_l(t)}} (1{-}d_{s, m}) {+} \sum_{\mathclap{m \in A_r(t)}} d_{s, m} {+} c_{t, a} {-} |A(t)| \leq \pi_{s, a},& \;
    \forall s, a, t \tag{\ref{eq:constraint-policy-follows-tree}} \\
    \sum_a \pi_{s, a} = 1,& \; \forall s \tag{\ref{eq:constraint-deterministic-policy}} \\
    x_{s, a} \leq M \pi_{s, a},& \; \forall s, a \tag{\ref{eq:constraint-frequency-follows-policy}}
\end{align}

\section{Results}
We present experiments comparing the performance of OMDTs with VIPER and dtcontrol. Viper uses imitation learning to extract a size-limited decision tree from a teacher policy and dtcontrol learns an unrestricted tree that exactly copies the teacher's behavior. To provide a fair comparison we have trained VIPER and dtcontrol with an unrestricted optimal teacher by first solving the MDP with value iteration and then extracting all Q values, both methods ran with default parameters. We also implemented and ran experiments on interpretable Differentiable Decision Trees~\cite{silva2020optimization} but excluded these models from our analysis as they did not outperform a random policy. The full code for OMDT and our experiments can be found on GitHub\footnote{\url{https://github.com/tudelft-cda-lab/OMDT}}. All of our experiments ran on a Linux machine with 16 Intel Xeon CPU cores and 72 GB of RAM total and used Gurobi 10.0.0 with default parameters. Each method ran on a single CPU core.

\begin{table*}[]
\centering
\setlength{\tabcolsep}{4.5pt}
\begin{tabular}{lrr|rrr|rr|r|rr}
\toprule
    & \multicolumn{1}{c}{} &  & \multicolumn{3}{c|}{normalized return} & \multicolumn{2}{c|}{MILP} & & \multicolumn{2}{c}{runtime (s)} \\
MDP & $|S|$ & $|A|$ & VIPER & \begin{tabular}[c]{@{}r@{}}OMDT\\ 5 mins.\end{tabular} & \begin{tabular}[c]{@{}r@{}}OMDT\\ 2 hrs.\end{tabular} & vars. & constrs. & trees & VIPER & \begin{tabular}[c]{@{}r@{}}OMDT\\ optimal\end{tabular} \\ \midrule
3d\_navigation & 125 & 6 & \textbf{1.00} ${\pm} .00$ & .81 ${\pm} .10$ & \textbf{1.00 ${\pm} .00$} & 2,528 & 7,890 & $10^{14}$ & 2,090 ${\pm} 55$ \:\: & 315 ${\pm} 89$ \:\: \\
blackjack & 533 & 2 & \textbf{1.00 ${\pm} .00$} & \textbf{1.00 ${\pm} .00$} & \textbf{1.00 ${\pm} .00$} & 6,187 & 14,406 & $10^{14}$ & 2,248 ${\pm} 27$ \:\: & 408 ${\pm} 85$ \:\: \\
frozenlake\_4x4 & 16 & 4 & .67 ${\pm} .00$ & \textbf{.96 ${\pm} .00$} & \textbf{.96 ${\pm} .00$} & 328 & 735 & $10^{10}$ & 74 ${\pm} 3$ \:\:\:\: & 2 ${\pm} 0$ \:\:\:\: \\
frozenlake\_8x8 & 64 & 4 & .83 ${\pm} .06$ & \textbf{.95 ${\pm} .00$} & \textbf{.95 ${\pm} .00$} & 1,104 & 2,895 & $10^{13}$ & 178 ${\pm} 5$ \:\:\:\: & 98 ${\pm} 30$ \:\: \\
frozenlake\_12x12 & 144 & 4 & .19 ${\pm} .09$ & .63 ${\pm} .03$ & \textbf{.68 ${\pm} .04$} & 2,360 & 6,495 & $10^{14}$ & 196 ${\pm} 38$ \:\: & \multicolumn{1}{c}{timeout} \\
inv. management & 101 & 100 & \textbf{1.00 ${\pm} .00$} & .37 ${\pm} .37$ & \textbf{1.00 ${\pm} .00$} & 22,414 & 91,824 & $10^{30}$ & 2,254 ${\pm} 86$ \:\: & 2,533 ${\pm} 540$ \\
sysadmin\_1 & 256 & 9 & .88 ${\pm} .01$ & .85 ${\pm} .01$ & \textbf{.92 ${\pm} .00$} & 6,584 & 23,055 & $10^{14}$ & 2,265 ${\pm} 37$ \:\: & \multicolumn{1}{c}{timeout} \\
sysadmin\_2 & 256 & 9 & \textbf{.59 ${\pm} .00$} & .23 ${\pm} .06$ & .58 ${\pm} .01$ & 6,584 & 23,055 & $10^{14}$ & 2,257 ${\pm} 7$ \:\:\:\: & \multicolumn{1}{c}{timeout} \\
sysadmin\_tree & 128 & 8 & .57 ${\pm} .04$ & .48 ${\pm} .07$ & \textbf{.70 ${\pm} .09$} & 3,106 & 10,383 & $10^{13}$ & 2,136 ${\pm} 72$ \:\: & \multicolumn{1}{c}{timeout} \\
tictactoe\_vs\_rand. & 2,424 & 9 & \textbf{.80 ${\pm} .01$} & -.06 ${\pm} .00$ & .43 ${\pm} .18$ & 61,239 & 218,175 & $10^{20}$ & 21 ${\pm} 3$ \:\:\:\: & \multicolumn{1}{c}{timeout} \\
tiger\_vs\_antelope & 626 & 5 & -.10 ${\pm} .02$ & -.17 ${\pm} .19$ & \textbf{.52 ${\pm} .03$} & 10,850 & 33,819 & $10^{15}$ & 490 ${\pm} 243$ & \multicolumn{1}{c}{timeout} \\
traffic\_intersec. & 361 & 2 & .98 ${\pm} .00$ & .99 ${\pm} .00$ & \textbf{1.00 ${\pm} .00$} & 4,127 & 9,762 & $10^{11}$ & 2,188 ${\pm} 121$ & 1,219 ${\pm} 177$ \\
xor & 200 & 2 & .34 ${\pm} .06$ & \textbf{1.00 ${\pm} .00$} & \textbf{1.00 ${\pm} .00$} & 5,016 & 5,415 & $10^{21}$ & 1,999 ${\pm} 123$ & 50 ${\pm} 0$ \:\:\:\: \\ \bottomrule
\end{tabular}
\caption{Comparison of depth 3 trees trained with VIPER and OMDT on 13 MDPs, experiments were repeated 3 times, means and standard errors are given. All runs were limited to 2 hours. OMDT solves some MDPs in 5 minutes but significantly improves when given 2 hours of runtime. While 2 hours are enough for OMDT to achieve greater or equal scores to VIPER in most MDPs, OMDT needs more time to outperform VIPER on the large tictactoe MDP. OMDT was able to identify the optimal size-limited tree and prove its optimality on 7 MDPs.}
\label{tab:omdt-vs-viper}
\end{table*}

\subsection{Environments}
For comparison we implemented 13 environments based on well-known MDPs from the literature, the sizes of these MDPs are given in Table \ref{tab:omdt-vs-viper}. All MDPs were pre-processed such that states that are unreachable from the initial states are removed. We briefly describe the environments below but refer to the appendix for complete descriptions.

In \textit{3d\_navigation} the agent controls a robot in a $5 {\times} 5 {\times} 5$ world and attempts to reach from start to finish with each voxel having a chance to make the robot disappear. \textit{blackjack} is a simplified version of the famous casino game where we assume an infinite-sized deck and only the actions `skip' or `hit'. \textit{frozenlake} is a grid world where the agent attempts to go from start to finish without falling into holes, actions are stochastic so the agent will not always move in the intended direction (e.g. the action `up' will only not send the agent `down'). \textit{inventory management} models a company that has to decide how many items to order to maximize profit while minimizing cost. \textit{system\_administrator} refers to a computer network where computers randomly crash and an administrator has to decide which computer to reboot. A crashed computer has an increased probability of crashing a neighboring computer. \textit{tictactoe\_vs\_random} is the well-known game of tic-tac-toe when played against an opponent that makes random moves. In \textit{tiger\_vs\_antelope} the agent attempts to catch an antelope that randomly jumps away from the tiger in a grid world. \textit{traffic\_intersection} describes a perpendicular intersection where traffic flows in at different rates and the operator decides when to switch the traffic lights. \textit{xor} is an MDP constructed with states randomly distributed on a plain, the agent gets 1 reward for taking the action according to an XOR function and -1 for a mistake. The XOR problem is notoriously difficult for greedy decision tree learning algorithms.

\subsection{Performance-Interpretability Trade-off}
\begin{figure}[tb]
    \centering
    \includegraphics[width=.99\linewidth]{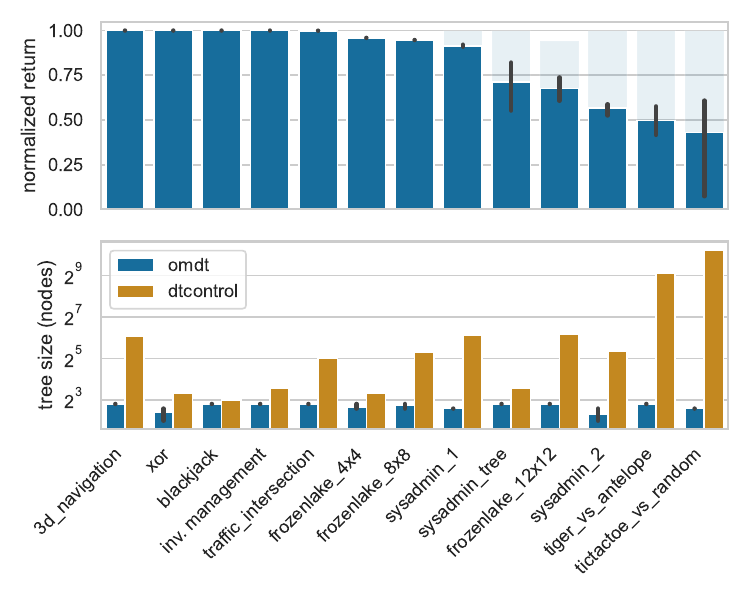}
    \caption{(top) Normalized return and bounds for OMDT trees of depth 3, optimal policies score 1 while uniform random policies score 0. (bottom) Log of tree sizes for OMDT (maximum depth 3) and dtcontrol. Dtcontrol always produces an optimal policy but the trees are orders of magnitude larger than OMDT.}
    \label{fig:performance-interpretability-trade-off}
\end{figure}

It is often assumed that there is a trade-off in the performance and interpretability of machine learning models~\cite{gunning2019darpa}, since interpretable models necessarily lack complexity but this assumption is not always true~\cite{rudin2019stop}. We aim to answer whether the performance-interpretability trade-off occurs in a variety of MDPs by training size-limited decision trees and comparing their performance to the optimal solutions that were not restricted in complexity. We visualize the normalized return of depth 3 OMDTs and unrestricted dtcontrol trees in Figure \ref{fig:performance-interpretability-trade-off}. Returns were normalized such that 0 corresponds to a random policy and 1 to an optimal one. Since small deterministic decision tree policies are limited in the number of distinct actions an optimal tree can perform worse than a random policy. Experiments were repeated 3 times and runs were limited to 2 hours. We consider an OMDT optimal when the relative gap between its objective and bound is proven to be less than 0.01\%.

While it is debatable what the precise size limits are for decision trees to be interpretable~\cite{lipton2018mythos} we use trees of depth 3 which implies that a tree has at most 8 leaves. Note that this also limits the number of distinct actions in the policy to 8. We find that in all environments, OMDTs of depth 3 improve on the performance of random policies, and in 8 out of 13 environments the policy gets close to optimal. Decision trees trained with dtcontrol always achieve the optimal normalized return of 1 since they exactly mimic the optimal policy. However, dtcontrol produces large trees that are not interpretable to humans. When run on 3d\_navigation for example, dtcontrol produces a tree of 68 decision nodes which is very complex for humans to understand. OMDT produces a tree of 7 decision nodes which performs equally well.

Overall, our results demonstrate that for small environments there is no performance-interpretability trade-off: simple policies represented by size-limited trees perform approximately as well as the unrestricted optimal policy.

\begin{figure*}[tb]
    \centering
    \centering
    \begin{subfigure}[b]{.32\linewidth}
        \centering
        \includegraphics[width=\textwidth]{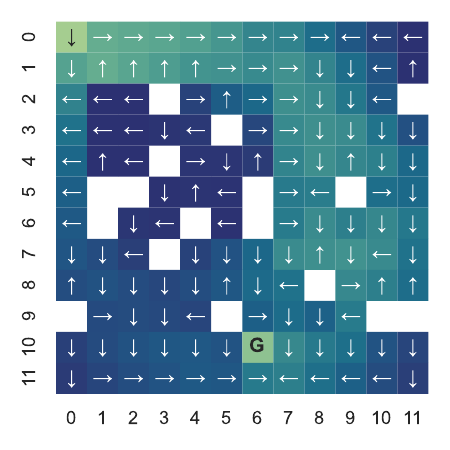}
        \caption{Optimal: 92\% success}
        \label{fig:blackjack-depth-1}
    \end{subfigure}
    \hfill
    \begin{subfigure}[b]{0.32\linewidth}
        \centering
        \includegraphics[width=\textwidth]{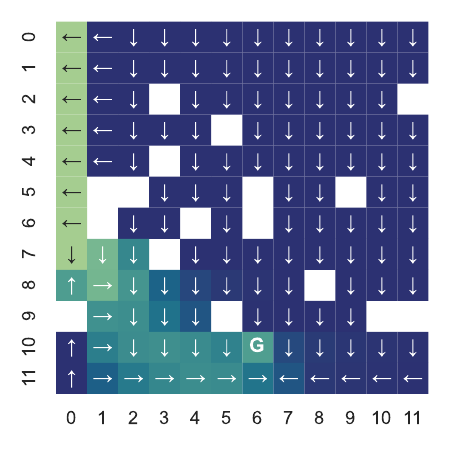}
        \caption{OMDT (depth 3): 66\% success}
        \label{fig:blackjack-depth-2}
    \end{subfigure}
    \hfill
    \begin{subfigure}[b]{.32\linewidth}
        \centering
        \includegraphics[width=\textwidth]{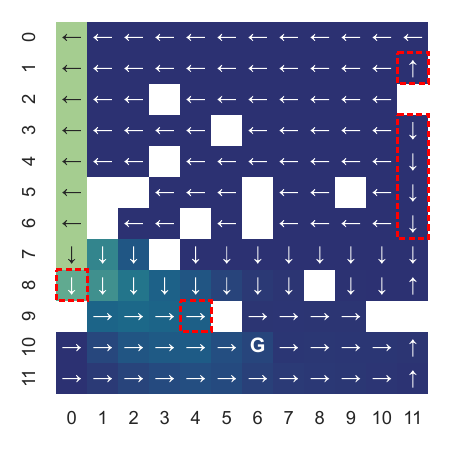}
        \caption{VIPER (depth 3): 11\% success}
        \label{fig:blackjack-depth-3}
    \end{subfigure}
    \caption{Paths taken on 10,000 Frozenlake\_12x12 runs. The agent starts at (0, 0) and attempts to reach the goal tile `G' while avoiding holes. Actions are indicated by arrows and are somewhat stochastic, i.e. an action of `up' will send the agent `left', `up', or `right' (but never down) with equal probability.
    VIPER fails to produce a good policy because it spends capacity of its tree mimicking parts of the complex teacher policy that its simple student policy will never reach. OMDT achieves a greater success rate by directly optimizing a simple policy.
    }
    \label{fig:frozenlake-12x12-policies}
\end{figure*}

\subsection{Direct Optimization versus Imitation Learning}
The above conclusion holds when the policy is learned to optimality under the constraint that it has to be a small tree, e.g., using OMDT. Techniques such as VIPER enforce the size constraint but aim to imitate the unrestricted optimal policy. We now show that this comes at a cost when the unrestricted policy is too complex to be represented using a small tree. 

VIPER trains trees by imitating high Q values of the optimal policies while OMDT directly maximizes expected return. In Table \ref{tab:omdt-vs-viper} we list the normalized return (0 for random policies, 1 for optimal policies) for VIPER and OMDT with respectively 5 minutes and 2 hours of runtime. After 5 minutes OMDT improves performance over random policies but often needs more time to improve over VIPER. After 2 hours OMDT's policies win on 11 out of 13 environments. For instances with large state space such as tictactoe\_vs\_random OMDT needs more than 2 hours to improve over VIPER.

\subsubsection{Shortcomings of Imitation Learning}
Overall, given sufficient runtime, OMDT produces better policies than VIPER. This cannot be easily solved by giving VIPER more runtime but is an inherent problem of imitation learning. 
To illustrate this, we investigate the results on the frozenlake MDPs as Table \ref{tab:omdt-vs-viper} demonstrates that imitation learning can perform far from optimal in these environments.

In theory, imitation learning performs optimally in the limit~\cite{ross2011reduction} but this result requires the student policy to be as expressive as the teacher policy. This is not the case for size-limited decision trees. When VIPER learns a policy for frozenlake\_12x12 it tries to imitate a complex policy using a small tree that cannot represent all teacher policy actions. This results in VIPER spending capacity of its decision tree on parts of the state space that will never be reached under its student policy. In Figure \ref{fig:frozenlake-12x12-policies} we visualize the paths that the agents took on 10,000 runs and indicate the policies with arrows. VIPER creates leaves that control action in the right section of the grid world (indicated in red). The optimal teacher policy often visits this section but the simple student does not. By directly optimizing a decision tree policy using OMDT, the policy spends its capacity on parts of the state space that it actually reaches. As a result, VIPER cannot prevent actions that send its agent into holes on the left part of the grid world (indicated in red). OMDT actively avoids these.

\subsection{Runtime}
Runtime for solving Mixed-Integer Linear Programming formulations scales worst-case exponentially which makes it important to understand how solvers operate on complex formulations such as OMDT. We solved OMDTs for a depth of 3 for a maximum of 2 hours and display the results in Table \ref{tab:omdt-vs-viper}. The table compares the runtimes of VIPER and solving OMDT to optimality. If the solver does not prove optimality within 2 hours we denote it as `timeout'. We also denote the number of possible decision tree policies computed as:
$
    |\mathcal{T}_B|^\text{possible splits} \times |\mathcal{T}_L|^{|A|}
$.
It provides an estimate of how many decision tree policies are possible and shows that enumerating trees with brute force is intractable.

OMDT solves a simple environment such as Frozenlake 4x4 (16 states, 4 actions) to optimality within 2 seconds but runtime grows for larger environments such as inventory management (101 states, 100 actions) which took on average 2533 seconds. VIPER needs roughly 2250 seconds of runtime for every MDP and runs significantly faster on some MDPs. This is because VIPER spends much time evaluating policies on the environment and some environments quickly reach terminal states which results in short episodes. While OMDT was able to prove optimality on only 7 out of 13 environments within 2 hours, OMDT finds good policies before this time on 12 out of 13 environments.

\section{Conclusion}
We propose OMDT, a Mixed-Integer Linear Programming formulation for training optimal size-limited decision trees for Markov Decision Processes. Our results show that for simple environments such as blackjack, we do not have to trade off interpretability for performance: OMDTs of depth 3 achieve near-optimal performance. On Frozenlake 12x12, OMDT outperforms VIPER by more than 100\%.

OMDT sets a foundation for extending supervised optimal decision tree learning techniques to the reinforcement learning setting. Still, OMDT requires a full specification of the Markov Decision Process. Imitation learning techniques such as VIPER can instead also learn from a simulation environment. Therefore, future work should focus on closing the gap between the theoretical bound supplied by OMDT and the practical performance achieved by algorithms that require only simulation access to optimize interpretable decision tree policies in reinforcement learning. Additionally, future work can incorporate factored MDPs into OMDT's formulation to scale up to larger state spaces.

\bibliographystyle{named}
\bibliography{ijcai23}

\clearpage

\appendix

\section{Environment Descriptions}
We describe the environments in additional detail below. All environments in the paper were solved with a discount rate of $\gamma = 0.99$. Such a value ensures that the solvers generate policies that have good performance even for environments that take hundreds of steps to solve. All MDPs were pre-processed such that states that are unreachable from the initial states are removed.

\subsection{3d\_navigation}
While policies in 2d scenarios are often easy to visualize and interpret policies in higher dimensions are harder to understand. Therefore we extend the 2d\_navigation problem from ICAP 2011 IPPC competition\footnote{\label{iccp}\url{http://users.cecs.anu.edu.au/~ssanner/IPPC_2011/}} to 3 dimensions. The agent controls a robot in a 5x5x5 world and attempts to reach from one corner to the other end with each voxel having a chance to make the robot disappear. The disappearing chances were independently sampled from a uniform distribution on [0, 1].

\subsection{blackjack}
The version of blackjack that we used\footnote{\url{https://gist.github.com/iiLaurens/ba9c479e71ee4ceef816ad50b87d9ebd}} is a slightly simplified version of the one that often appears in casinos. Specifically, we make the following assumptions:
\begin{itemize}
    \item There is no doubling down and splitting action.
    \item The player does not have knowledge about other players at the board.
    \item Aces always account for a value of 11.
    \item Cards are drawn from an infinite-sized deck, i.e. drawn with replacement.
\end{itemize}

The player always starts in the state where no cards are dealt, the dealer receives a random card, followed by the player receiving a random card. The player can 'Draw' cards until they cross the total of 21 or until they 'Skip'. When that happens the dealer receives cards until their total is over 17. In the end, the reward is computed as follows:
\begin{itemize}
    \item -1 if the player has a lower total than the dealer or if the player went over 21.
    \item 0 if the player and dealer end up with the same total.
    \item +1 if the player has a higher total than the dealer or if the dealer went over 21.
    \item +1.5 if the player gets to exactly 21
\end{itemize}

\subsection{frozenlake}
Frozenlake is a maze-like game played on a 2D grid where the player attempts to move from the start state (S) to the goal state (G) without falling into holes (H). The player walks over frozen tiles (F) that make the actions stochastic:
\begin{itemize}
    \item The action 'Up' sends a player in the directions left, up, or right with probability $1/3$ each.
    \item The action 'Right' sends a player in the directions up, right, or down with probability $1/3$ each.
    \item The action 'Down' sends a player in the directions right, down, or left with probability $1/3$ each.
    \item The action 'Left' sends a player in the directions down, left, or up with probability $1/3$ each.
\end{itemize}

We used the default 4x4 and 8x8 maps as defined by the OpenAI gym\footnote{\url{https://github.com/openai/gym/blob/master/gym/envs/toy_text/frozen_lake.py}} and defined a new map for 12x12. The maps are given below (in the same format that the environment uses):

{%
\begin{minipage}{.21\linewidth}
\begin{center}
4x4
\end{center}
\begin{Verbatim}[commandchars=\\\{\}]
"\textcolor{blue}{S}FFF",
"F\textcolor{red}{H}F\textcolor{red}{H}",
"FFF\textcolor{red}{H}",
"\textcolor{red}{H}FF\textcolor{green}{G}",
\end{Verbatim}
\end{minipage}
\begin{minipage}{.31\linewidth}
\begin{center}
8x8
\end{center}
\begin{Verbatim}[commandchars=\\\{\}]
"\textcolor{blue}{S}FFFFFFF",
"FFFFFFFF",
"FFF\textcolor{red}{H}FFFF",
"FFFFF\textcolor{red}{H}FF",
"FFF\textcolor{red}{H}FFFF",
"F\textcolor{red}{H}\textcolor{red}{H}FFF\textcolor{red}{H}F",
"F\textcolor{red}{H}FF\textcolor{red}{H}F\textcolor{red}{H}F",
"FFF\textcolor{red}{H}FFF\textcolor{green}{G}",
\end{Verbatim}
\end{minipage}
\begin{minipage}{.41\linewidth}
\begin{center}
12x12
\end{center}
\begin{Verbatim}[commandchars=\\\{\}]
"\textcolor{blue}{S}FFFFFFFFFFF",
"FFFFFFFFFFFF",
"FFF\textcolor{red}{H}FFFFFFF\textcolor{red}{H}",
"FFFFF\textcolor{red}{H}FFFFFF",
"FFF\textcolor{red}{H}FFFFFFFF",
"F\textcolor{red}{H}\textcolor{red}{H}FFF\textcolor{red}{H}FF\textcolor{red}{H}FF",
"F\textcolor{red}{H}FF\textcolor{red}{H}F\textcolor{red}{H}FFFFF",
"FFF\textcolor{red}{H}FFFFFFFF",
"FFFFFFFF\textcolor{red}{H}FFF",
"\textcolor{red}{H}FFFF\textcolor{red}{H}FFFF\textcolor{red}{H}\textcolor{red}{H}",
"FFFFFF\textcolor{green}{G}FFFFF",
"FFFFFFFFFFFF",
\end{Verbatim}
\end{minipage}}

\subsection{inventory management}
In inventory management, the player is the owner of a shop that stocks items and sells them to customers. Our environment is adapted from the implementation by Paul Hendricks\footnote{\url{https://github.com/paulhendricks/gym-inventory/tree/master}}. Every day the player can choose `Don't buy' or `Buy x' items. We used the following parameters for the problem:
\begin{itemize}
    \item The maximum inventory size is 100 (this creates 101 states since the inventory can have 0 up to and including 100 items)
    \item Ordering any items comes with a fixed cost of -10
    \item Ordering items costs -2 per item
    \item Holding an item in storage costs -1 per item
    \item Selling an item grants the player +4
    \item The number of customers follows a Poisson distribution with expectation $\lambda = 15$ (customers per day)
\end{itemize}

\subsection{system administrator}
\begin{figure}[tb]
    \centering
    \centering
    \begin{subfigure}[b]{.32\linewidth}
        \centering
        \includegraphics[width=\textwidth]{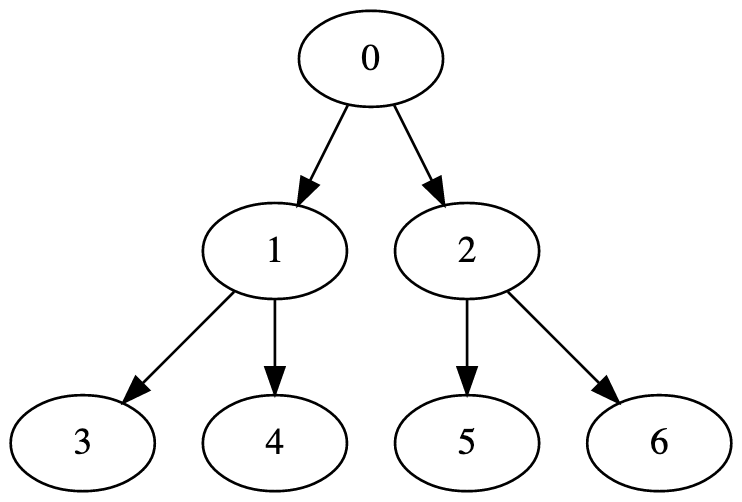}
        \caption{tree}
    \end{subfigure}
    \hfill
    \begin{subfigure}[b]{0.23\linewidth}
        \centering
        \includegraphics[width=\textwidth]{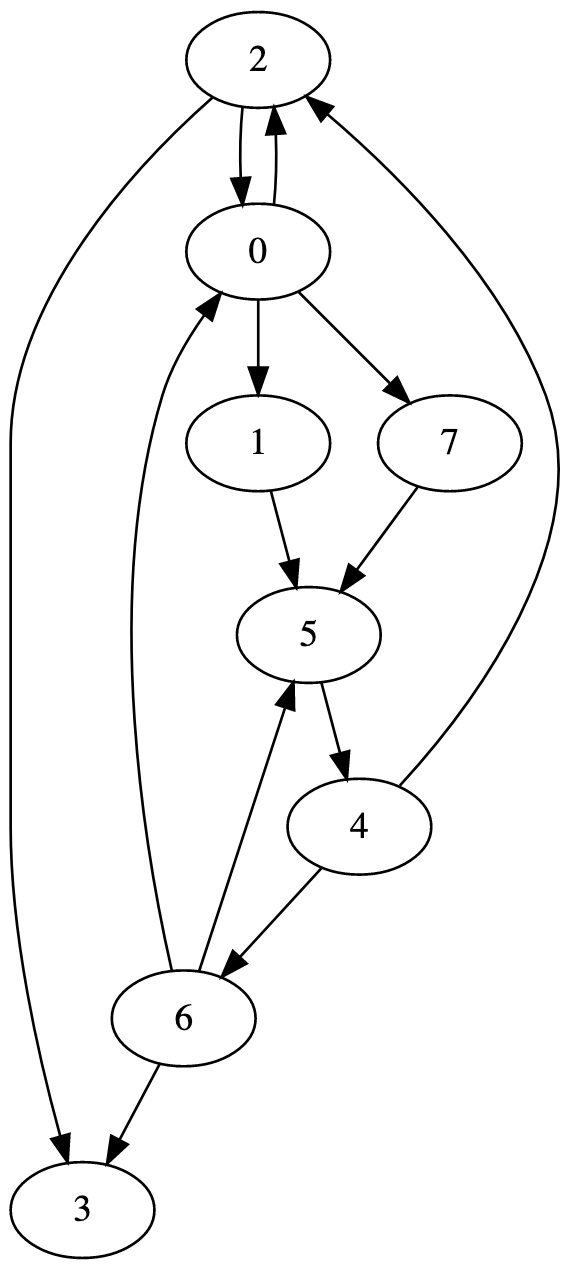}
        \caption{random 1}
    \end{subfigure}
    \hfill
    \begin{subfigure}[b]{.35\linewidth}
        \centering
        \includegraphics[width=\textwidth]{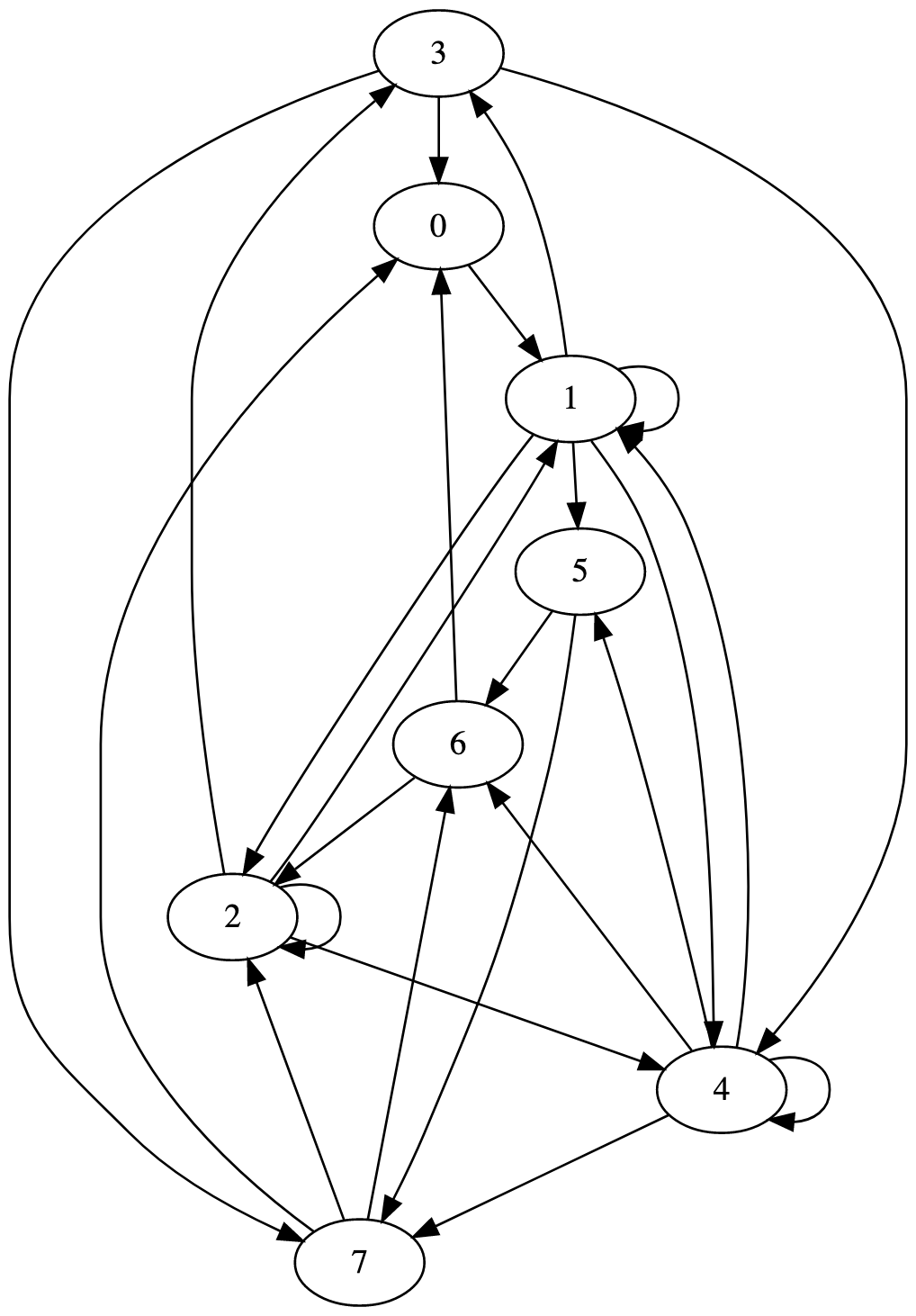}
        \caption{random 2}
    \end{subfigure}
    \caption{
    System administrator topologies.
    }
    \label{fig:sysadmin-topologies}
\end{figure}

The system administrator task refers to a computer network where computers randomly crash and an administrator has to decide which computer to spend time on rebooting. This environment is based on one of the MDPs of the ICAPS 2011 IPPC competition\textsuperscript{\ref{iccp}}. When a computer crashes it has an increased probability in following time steps to crash a neighboring computer. The topology of the network can be varied to create MDPs with various properties. The 3 topologies considered in this work are visualized in Figure \ref{fig:sysadmin-topologies}.

The system administrator is then allowed to reboot one machine at every time step or wait. Rebooting a computer is penalized with a reward of -0.45 but will always fix the crashed computer. At each time step, every computer that has not crashed will reward the system administrator with +1. We refer to a computer being on (1) or crashed (0) as the `status', the probability of being on in the next step is then given by:
\begin{equation*}
    \text{ratio\_on\_neighbors} \times (0.05 + 0.9 \times \text{status})
\end{equation*}

\subsection{tictactoe\_vs\_random}
The well-known tic-tac-toe game is played on a 3x3 grid where players take turns putting a cross/circle in an empty square. The player that manages to put 3 of their symbols in a row/column / diagonal wins the game. While 2 player zero-sum games are not MDPs, they become MDPs when we fix the strategy of one of the players. In this case, the opponent plays moves uniformly at random. It is worth noting that this turns the game that is normally deterministic into a stochastic environment. The player gets to start the game, i.e. they play with the crosses.

\subsection{tiger\_vs\_antelope}
The tiger vs antelope game is played on a 5x5 grid where the agent controls the tiger and tries to catch an antelope. Every turn the agent can move `up', `down', `left', or `right', and if the antelope is not caught it responds by jumping in any direction away from the player uniformly at random.

\subsection{traffic\_intersection}
The traffic intersection scenario that we considered describes a perpendicular intersection where traffic flows in at different rates and the operator decides when to switch the traffic lights. At each side of the intersection, 5 cars can fit in a row, and from the moment the traffic light switches we count up to 5 time steps in which one side of the intersection is waiting. Cars arrive with a probability of 0.1 per time step on one side of the intersection and 0.5 on the other side.

The operator gets rewarded +1 for each car making it through the intersection and gets penalized for flipping the color of the traffic signs and for waiting cars. Flipping the color of the traffic lights incurs a reward of -2. For each waiting car, the traffic operator gets a reward of $-0.1 \times 2^{\text{wait\_time}}$. 

\subsection{xor}
The XOR problem is notoriously difficult for greedy decision tree learning algorithms, we generate an MDP that mimics the structure of the supervised learning problem. Specifically, we draw 200 samples in 2 dimensions uniformly at random. The agent gets 1 reward for taking the action $a$ according to an XOR function and -1 for a mistake:
\begin{equation*}
R(x, y, a) = \begin{cases}
    1 & \text{round}(x) + \text{round}(y) \mod 2 = a\\
    -1 & \text{round}(x) + \text{round}(y) \mod 2 = a
\end{cases}
\end{equation*}
Each action sends the agent to a state uniformly at random.

\section{Linearizing Implications of Conjunctions}
In the main text, we constructed one of OMDT's constraints using the fact that:
\begin{align*}
    x_1 \land x_2 \land ... \land x_n {\implies} y
    \equiv x_1 + x_2 + ... + x_n - n + 1 \leq y 
\end{align*}
We give a short proof of this statement. First, we rewrite the logical formula into conjunctive normal form. In this case a single disjunctive clause.
\begin{align*}
    x_1 \land x_2 \land ... \land x_n \implies y \\
    \equiv \neg(x_1 \land x_2 \land ... \land x_n) \lor y \\
    \equiv \neg x_1 \lor \neg x_2 \lor ... \lor \neg x_n \lor y
\end{align*}
A disjunctive constraint $a \lor b \lor c$ can be trivially expressed as the linear constraint $a + b + c \geq 1$. Intuitively any variable $a$, $b$, or $c$ needs to be true. Now we rewrite to linear constraints:
\begin{align*}
    \neg x_1 \lor \neg x_2 \lor ... \lor \neg x_n \lor y \\
    \equiv (1 - x_1) + (1 - x_2) + ... + (1 - x_n) + y \geq 1 \\
    \equiv x_1 + x_2 + ... + x_n - n + 1 \leq y 
\end{align*}

\section{Big-M Value}
For the constraints that force $x_{s,a}$ according to the policy $\pi_{s,a}$ (Equation \ref{eq:constraint-frequency-follows-policy}) we used a big-M formulation with $M = \frac{1}{1 - \gamma}$ as an upper bound on $x_{s,a}$. To prove this upper bound we construct an MDP in which we maximize $x_{s,a}$ and then show that this value equals our bound. Consider an MDP with one state $s^*$ and one action $a^*$ such that the agent always starts in that state ($p_0(s^*) = 1$) and action $a^*$ will always return to $s^*$ with probability $P_{s^*, s^*, a^*} = 1$. Clearly, this maximizes the frequency $x_{s^*, a^*}$. Substituting into Equation \ref{eq:dual-constraint} we find: 
\begin{equation}
\sum_a x_{s, a} = p_0(s) + \sum_{s'} \sum_a \gamma P_{s', s, a} x_{s', a} \tag{\ref{eq:dual-constraint}}
\end{equation}
\begin{equation*}
x_{s^*, a^*} = p_0(s^*) + \gamma x_{s^*, a^*} = \frac{1}{1 - \gamma}
\end{equation*}

\section{Performance at Varying Depths}
In the main text, we discussed the performance of trees at a depth of 3 since these trees are still easily interpretable but more powerful than trees of depth 1 or 2. Normalized returns after 2 hours of runtime for varying depths are presented in Table~\ref{tab:normalized-returns-depths}, Table~\ref{tab:optimality-runtime-depths} shows the runtime needed to prove the optimality of the solutions for varying depths.

\begin{table*}[p]
\centering
\begin{tabular}{l|rr|rr|rr|rr}
\toprule
 & \multicolumn{2}{c|}{depth 1} & \multicolumn{2}{c|}{depth 2} & \multicolumn{2}{c|}{depth 3} & \multicolumn{2}{c}{depth 4} \\
MDP & OMDT & VIPER & OMDT & VIPER & OMDT & VIPER & OMDT & VIPER \\ \midrule
3d\_navigation &
\textbf{-.00} \tiny $\pm$ .00 &
\textbf{-.00} \tiny $\pm$ .00 &
\textbf{.06} \tiny $\pm$ .00 &
.01 \tiny $\pm$ .00 &
\textbf{1.00} \tiny $\pm$ .00 &
\textbf{1.00} \tiny $\pm$ .00 &
\textbf{1.00} \tiny $\pm$ .00 &
\textbf{1.00} \tiny $\pm$ .00 \\
blackjack &
\textbf{.98} \tiny $\pm$ .00 &
\textbf{.98} \tiny $\pm$ .00 &
\textbf{1.00} \tiny $\pm$ .00 &
.98 \tiny $\pm$ .00 &
\textbf{1.00} \tiny $\pm$ .00 &
\textbf{1.00} \tiny $\pm$ .00 &
\textbf{1.00} \tiny $\pm$ .00 &
\textbf{1.00} \tiny $\pm$ .00 \\
frozenlake\_4x4 &
\textbf{.19} \tiny $\pm$ .00 &
.04 \tiny $\pm$ .06 &
\textbf{.67} \tiny $\pm$ .00 &
.46 \tiny $\pm$ .00 &
\textbf{.96} \tiny $\pm$ .00 &
.67 \tiny $\pm$ .00 &
\textbf{1.00} \tiny $\pm$ .00 &
\textbf{1.00} \tiny $\pm$ .00 \\
frozenlake\_8x8 &
\textbf{.74} \tiny $\pm$ .00 &
.72 \tiny $\pm$ .01 &
\textbf{.93} \tiny $\pm$ .00 &
.90 \tiny $\pm$ .02 &
\textbf{.95} \tiny $\pm$ .00 &
.83 \tiny $\pm$ .06 &
\textbf{.98} \tiny $\pm$ .00 &
.95 \tiny $\pm$ .00 \\
frozenlake\_12x12 &
\textbf{.06} \tiny $\pm$ .00 &
.04 \tiny $\pm$ .02 &
\textbf{.34} \tiny $\pm$ .00 &
.30 \tiny $\pm$ .00 &
\textbf{.68} \tiny $\pm$ .04 &
.19 \tiny $\pm$ .09 &
\textbf{.81} \tiny $\pm$ .01 &
.19 \tiny $\pm$ .09 \\
inv. management &
\textbf{.99} \tiny $\pm$ .00 &
.98 \tiny $\pm$ .00 &
\textbf{1.00} \tiny $\pm$ .00 &
.99 \tiny $\pm$ .00 &
\textbf{1.00} \tiny $\pm$ .00 &
\textbf{1.00} \tiny $\pm$ .00 &
\textbf{1.00} \tiny $\pm$ .00 &
\textbf{1.00} \tiny $\pm$ .00 \\
sysadmin\_1 &
\textbf{.84} \tiny $\pm$ .00 &
.83 \tiny $\pm$ .00 &
\textbf{.89} \tiny $\pm$ .00 &
.86 \tiny $\pm$ .03 &
\textbf{.92} \tiny $\pm$ .00 &
.88 \tiny $\pm$ .01 &
.91 \tiny $\pm$ .00 &
\textbf{.92} \tiny $\pm$ .01 \\
sysadmin\_2 &
\textbf{.27} \tiny $\pm$ .00 &
\textbf{.27} \tiny $\pm$ .00 &
\textbf{.55} \tiny $\pm$ .00 &
\textbf{.55} \tiny $\pm$ .00 &
.57 \tiny $\pm$ .02 &
\textbf{.59} \tiny $\pm$ .00 &
.67 \tiny $\pm$ .04 &
\textbf{.72} \tiny $\pm$ .00 \\
sysadmin\_tree &
\textbf{-.03} \tiny $\pm$ .00 &
\textbf{-.03} \tiny $\pm$ .00 &
\textbf{.37} \tiny $\pm$ .00 &
.26 \tiny $\pm$ .00 &
\textbf{.71} \tiny $\pm$ .08 &
.57 \tiny $\pm$ .04 &
\textbf{.86} \tiny $\pm$ .04 &
\textbf{.86} \tiny $\pm$ .00 \\
tictactoe\_vs\_random &
\textbf{-.06} \tiny $\pm$ .00 &
\textbf{-.06} \tiny $\pm$ .00 &
\textbf{.61} \tiny $\pm$ .00 &
\textbf{.61} \tiny $\pm$ .00 &
.43 \tiny $\pm$ .18 &
\textbf{.80} \tiny $\pm$ .01 &
.68 \tiny $\pm$ .10 &
\textbf{.83} \tiny $\pm$ .00 \\
tiger\_vs\_antelope &
\textbf{-.36} \tiny $\pm$ .00 &
-.64 \tiny $\pm$ .06 &
\textbf{.19} \tiny $\pm$ .05 &
-.31 \tiny $\pm$ .17 &
\textbf{.50} \tiny $\pm$ .05 &
-.10 \tiny $\pm$ .02 &
\textbf{.64} \tiny $\pm$ .03 &
.23 \tiny $\pm$ .10 \\
traffic\_intersection &
\textbf{.50} \tiny $\pm$ .00 &
.42 \tiny $\pm$ .00 &
\textbf{.98} \tiny $\pm$ .00 &
\textbf{.98} \tiny $\pm$ .00 &
\textbf{1.00} \tiny $\pm$ .00 &
.98 \tiny $\pm$ .00 &
\textbf{1.00} \tiny $\pm$ .00 &
\textbf{1.00} \tiny $\pm$ .00 \\
xor &
\textbf{.15} \tiny $\pm$ .00 &
.05 \tiny $\pm$ .00 &
\textbf{1.00} \tiny $\pm$ .00 &
.13 \tiny $\pm$ .00 &
\textbf{1.00} \tiny $\pm$ .00 &
.34 \tiny $\pm$ .06 &
\textbf{1.00} \tiny $\pm$ .00 &
.90 \tiny $\pm$ .03 \\ \bottomrule
\end{tabular}
\caption{Normalized returns of OMDT and VIPER when varying the maximum depth of the decision tree, the runtime was limited to 2 hours, and experiments were repeated with 3 random seeds. For depth 3 OMDT finds optimal or high-quality trees within 2 hours while for depth 4 OMDT will need more runtime to improve on the scores produced by VIPER's trees in 3 environments.}
\label{tab:normalized-returns-depths}
\end{table*}

\begin{table*}[p]
\centering
\setlength{\tabcolsep}{10pt}
\begin{tabular}{l|rrrr}
\toprule
MDP & depth 1 & depth 2 & depth 3 & depth 4 \\ \midrule
3d\_navigation & 9 \tiny $\pm 0$ & 258 \tiny $\pm 27$ & 315 \tiny $\pm 89$ & 223 \tiny $\pm 25$ \\
blackjack & 36 \tiny $\pm 5$ & 72 \tiny $\pm 7$ & 408 \tiny $\pm 85$ & 598 \tiny $\pm 99$ \\
frozenlake\_4x4 & 1 \tiny $\pm 0$ & 1 \tiny $\pm 0$ & 2 \tiny $\pm 0$ & 2 \tiny $\pm 0$ \\
frozenlake\_8x8 & 2 \tiny $\pm 0$ & 9 \tiny $\pm 2$ & 98 \tiny $\pm 30$ & 3134 \tiny $\pm 350$ \\
frozenlake\_12x12 & 10 \tiny $\pm 2$ & 349 \tiny $\pm 57$ & \multicolumn{1}{c}{timeout} & \multicolumn{1}{c}{timeout} \\
inv. management & 453 \tiny $\pm 42$ & 6597 \tiny $\pm 507$ & 2533 \tiny $\pm 540$ & 3548 \tiny $\pm 732$ \\
sysadmin\_1 & 129 \tiny $\pm 10$ & 5059 \tiny $\pm 542$ & \multicolumn{1}{c}{timeout} & \multicolumn{1}{c}{timeout} \\
sysadmin\_2 & 488 \tiny $\pm 53$ & \multicolumn{1}{c}{timeout} & \multicolumn{1}{c}{timeout} & \multicolumn{1}{c}{timeout} \\
sysadmin\_tree & 74 \tiny $\pm 10$ & 4547 \tiny $\pm 767$ & \multicolumn{1}{c}{timeout} & \multicolumn{1}{c}{timeout} \\
tictactoe\_vs\_random & 2989 \tiny $\pm 61$ & \multicolumn{1}{c}{timeout} & \multicolumn{1}{c}{timeout} & \multicolumn{1}{c}{timeout} \\
tiger\_vs\_antelope & 1695 \tiny $\pm 210$ & \multicolumn{1}{c}{timeout} & \multicolumn{1}{c}{timeout} & \multicolumn{1}{c}{timeout} \\
traffic\_intersection & 33 \tiny $\pm 1$ & 104 \tiny $\pm 28$ & 1219 \tiny $\pm 177$ & \textit{5595} \;\;\;\; \\
xor & 113 \tiny $\pm 10$ & 43 \tiny $\pm 1$ & 50 \tiny $\pm 0$ & 98 \tiny $\pm 27$ \\ \bottomrule
\end{tabular}
\caption{Runtimes until OMDT proves optimality when varying the depth of the trees, the runtime was limited to 2 hours, and experiments repeated with 3 random seeds. Proving optimality for deeper trees usually takes more runtime but since a deeper tree allows the tree's objective value to be closer to the bound runtime can reduce e.g. in \textit{3d\_navigation} depth 4. OMDT only proved optimality for \textit{traffic\_intersection} depth 4 in one of the runs.}
\label{tab:optimality-runtime-depths}
\end{table*}

\section{Size-Limited Policies}
\begin{figure*}[p]
    \centering
    \begin{subfigure}[b]{.32\linewidth}
        \centering
        \includegraphics[width=\textwidth]{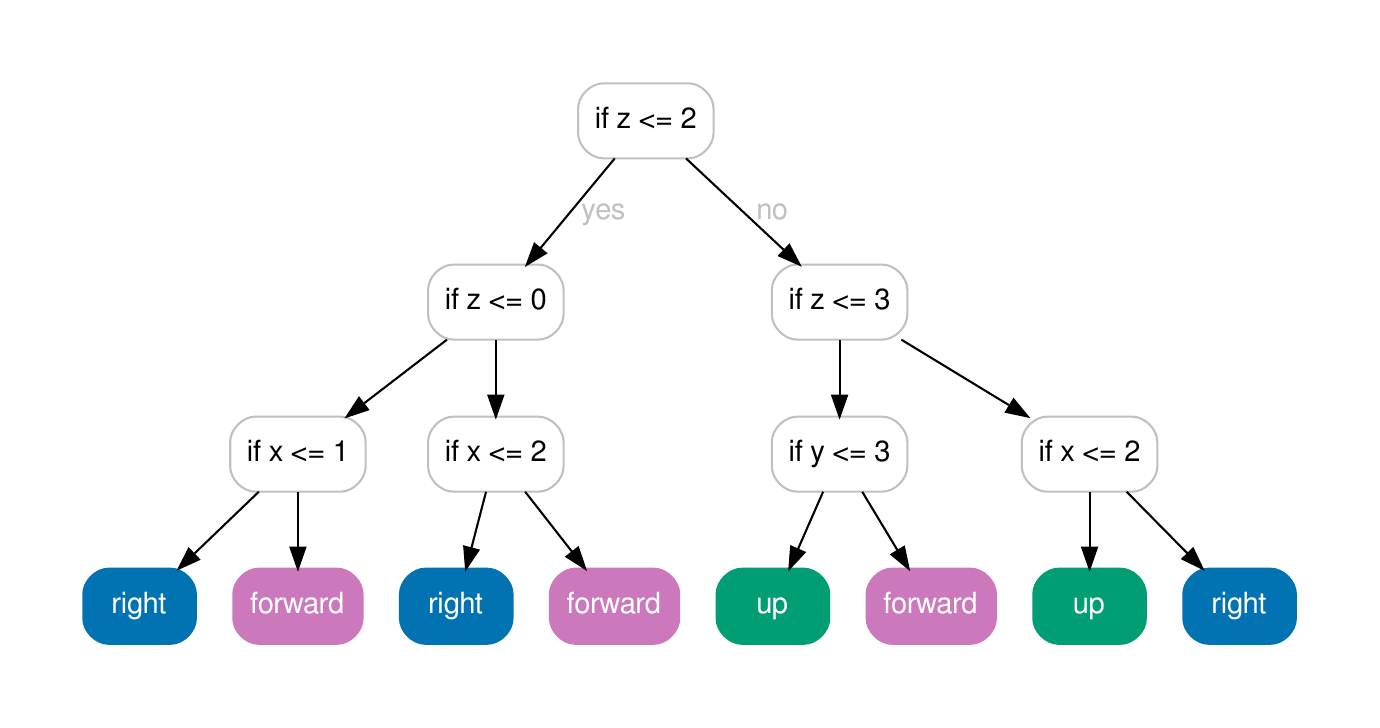}
        \caption{3d\_navigation}
    \end{subfigure}
    \hfill
    \begin{subfigure}[b]{0.32\linewidth}
        \centering
        \includegraphics[width=\textwidth]{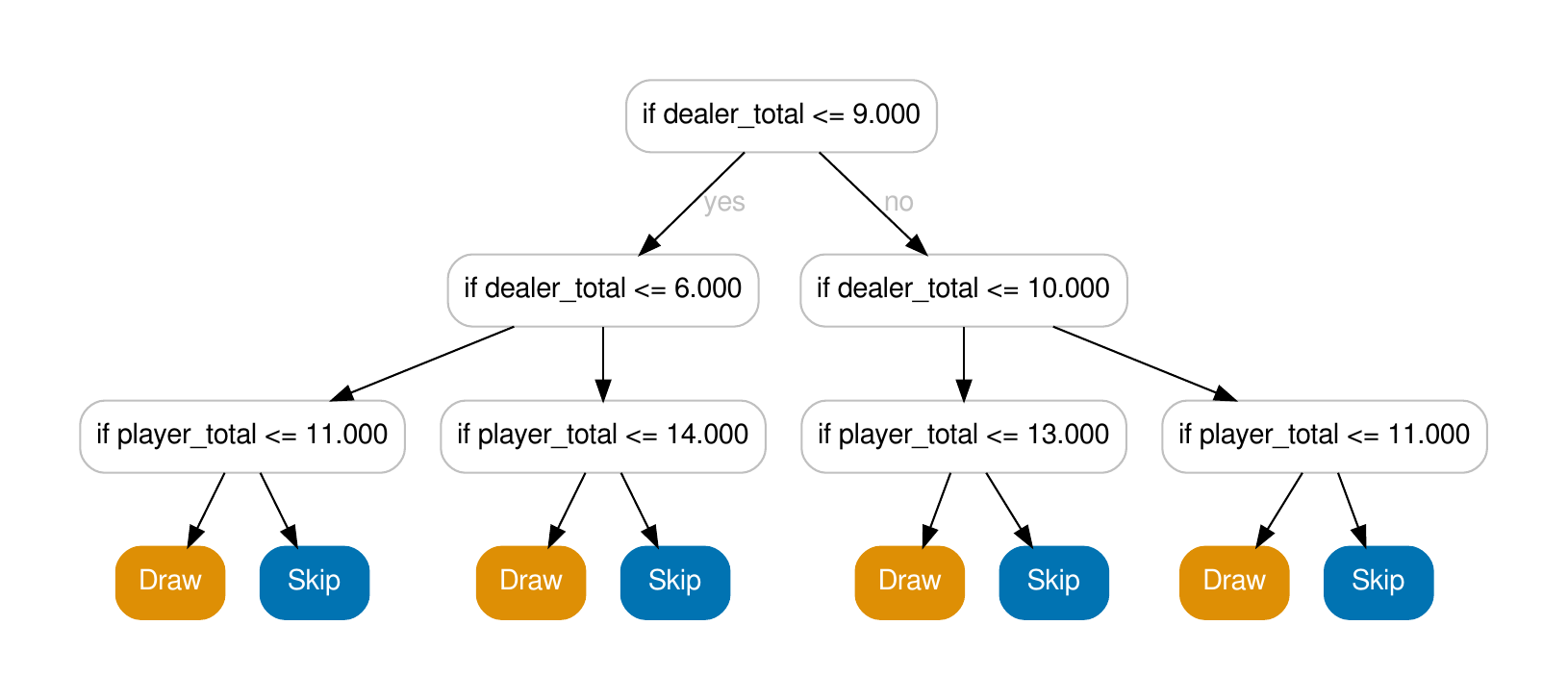}
        \caption{blackjack}
    \end{subfigure}
    \hfill
    \begin{subfigure}[b]{.32\linewidth}
        \centering
        \includegraphics[width=\textwidth]{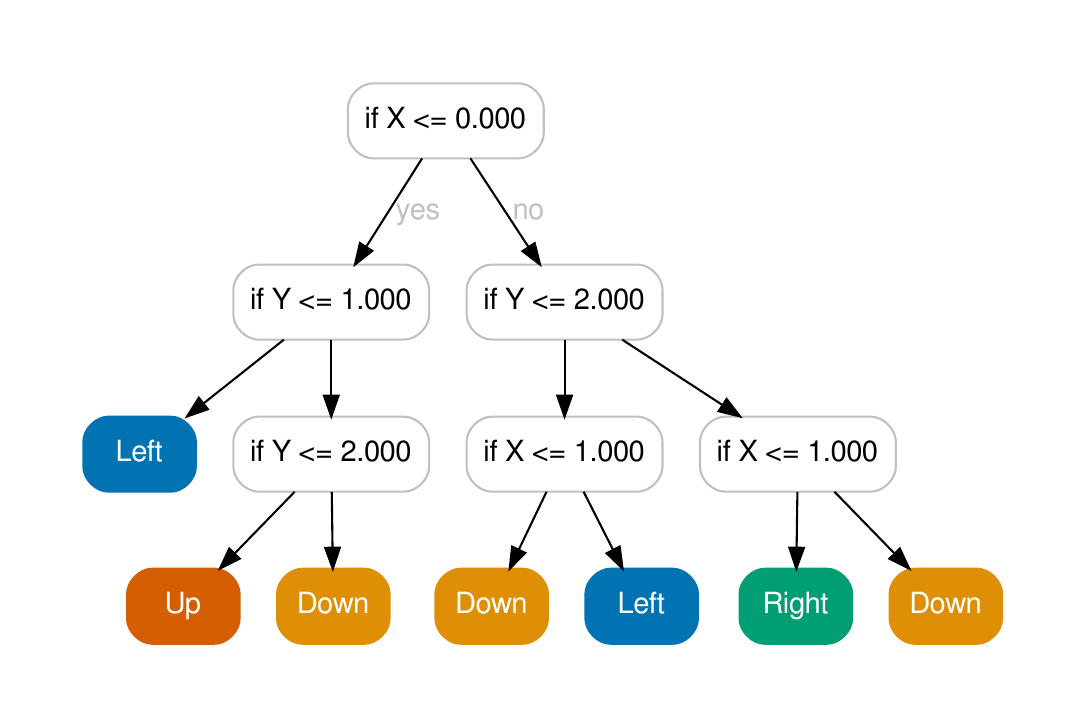}
        \caption{frozenlake\_4x4}
    \end{subfigure}
    \newline
    \begin{subfigure}[b]{.32\linewidth}
        \centering
        \includegraphics[width=\textwidth]{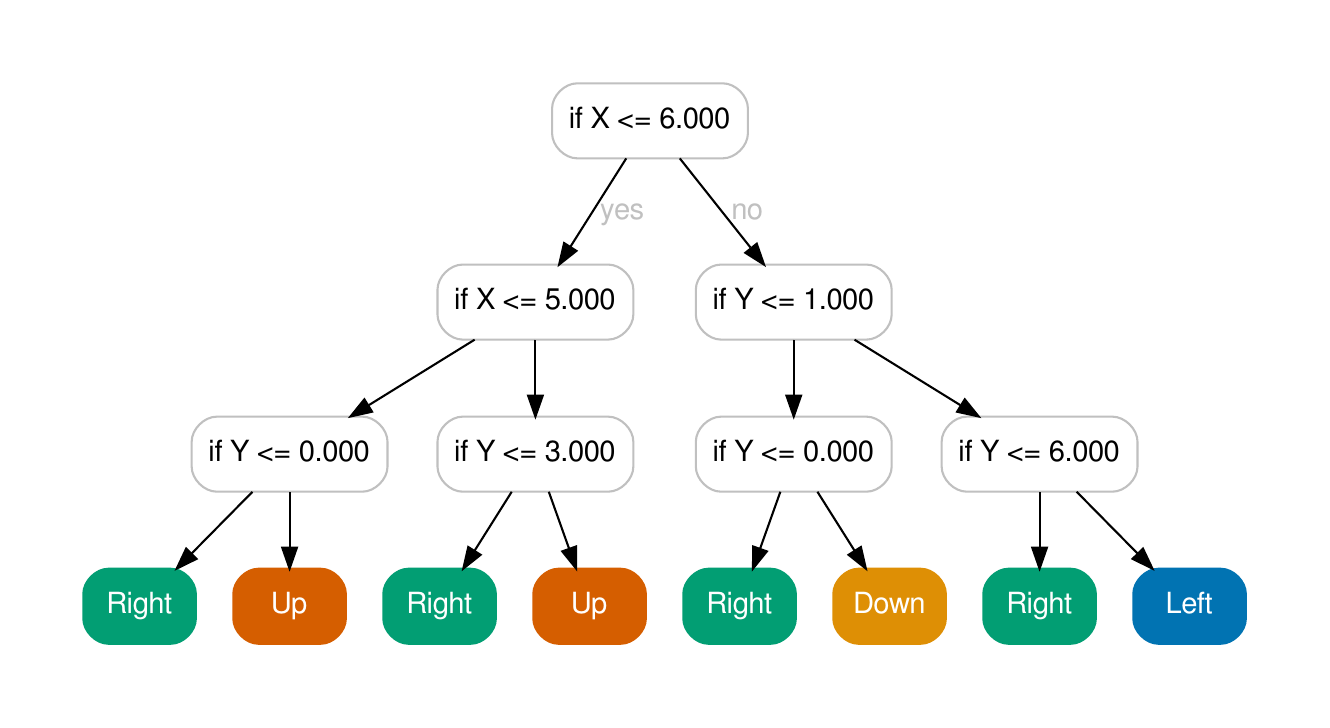}
        \caption{frozenlake\_8x8}
    \end{subfigure}
    \hfill
    \begin{subfigure}[b]{0.32\linewidth}
        \centering
        \includegraphics[width=\textwidth]{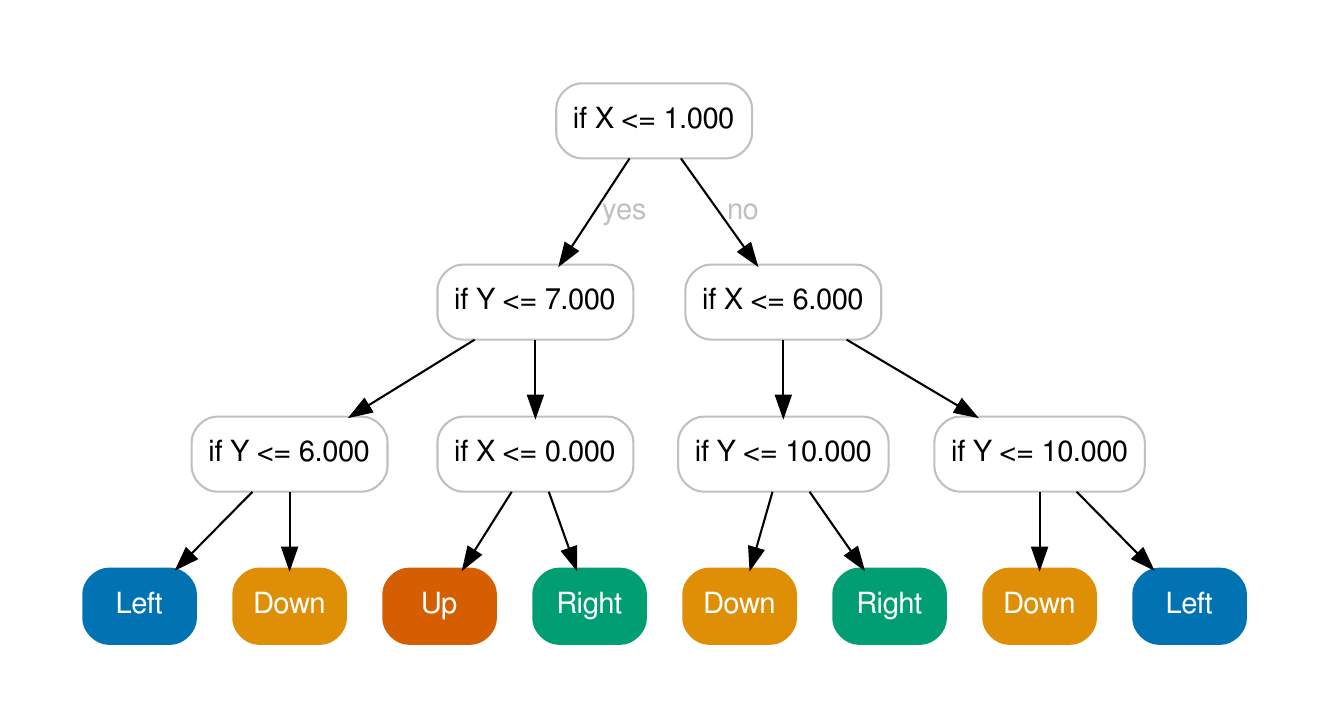}
        \caption{frozenlake\_12x12}
    \end{subfigure}
    \hfill
    \begin{subfigure}[b]{.32\linewidth}
        \centering
        \includegraphics[width=\textwidth]{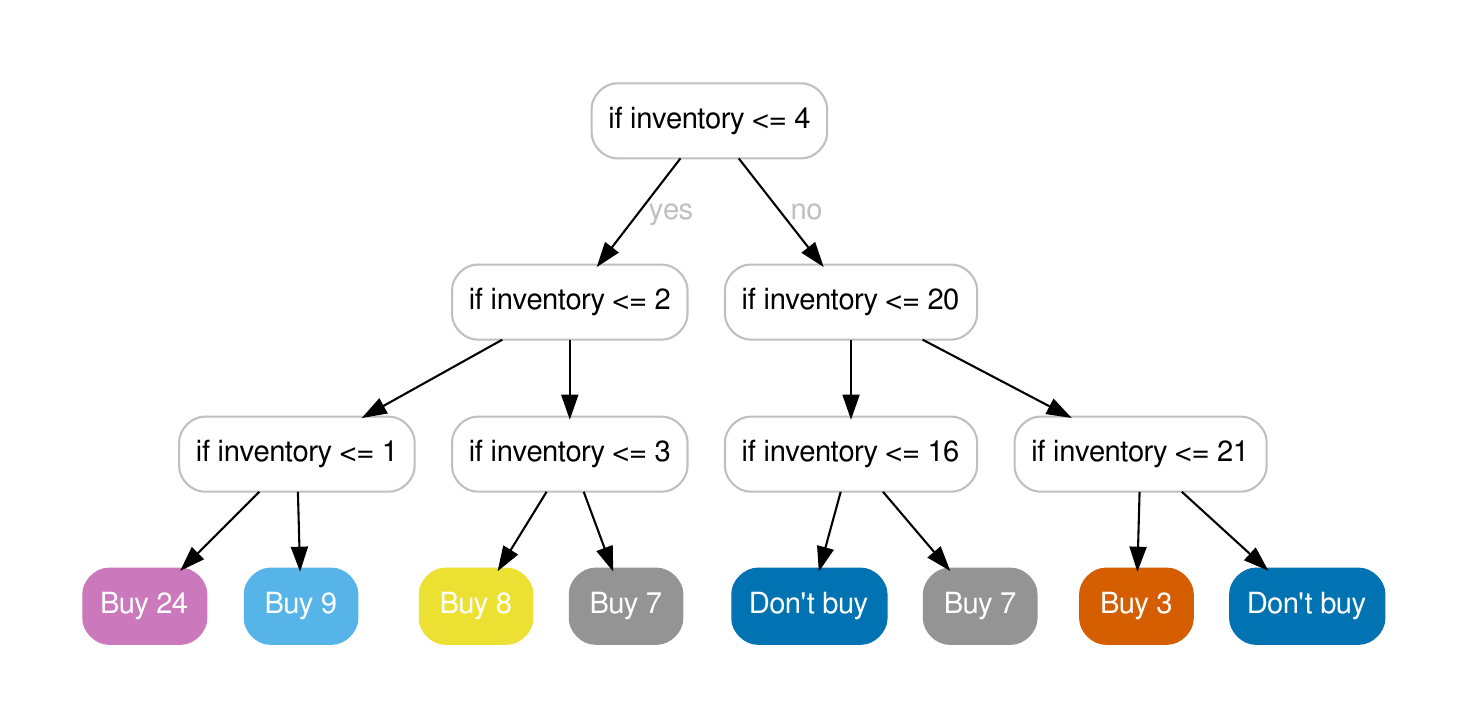}
        \caption{inventory\_management}
    \end{subfigure}
    \newline
    \begin{subfigure}[b]{.32\linewidth}
        \centering
        \includegraphics[width=\textwidth]{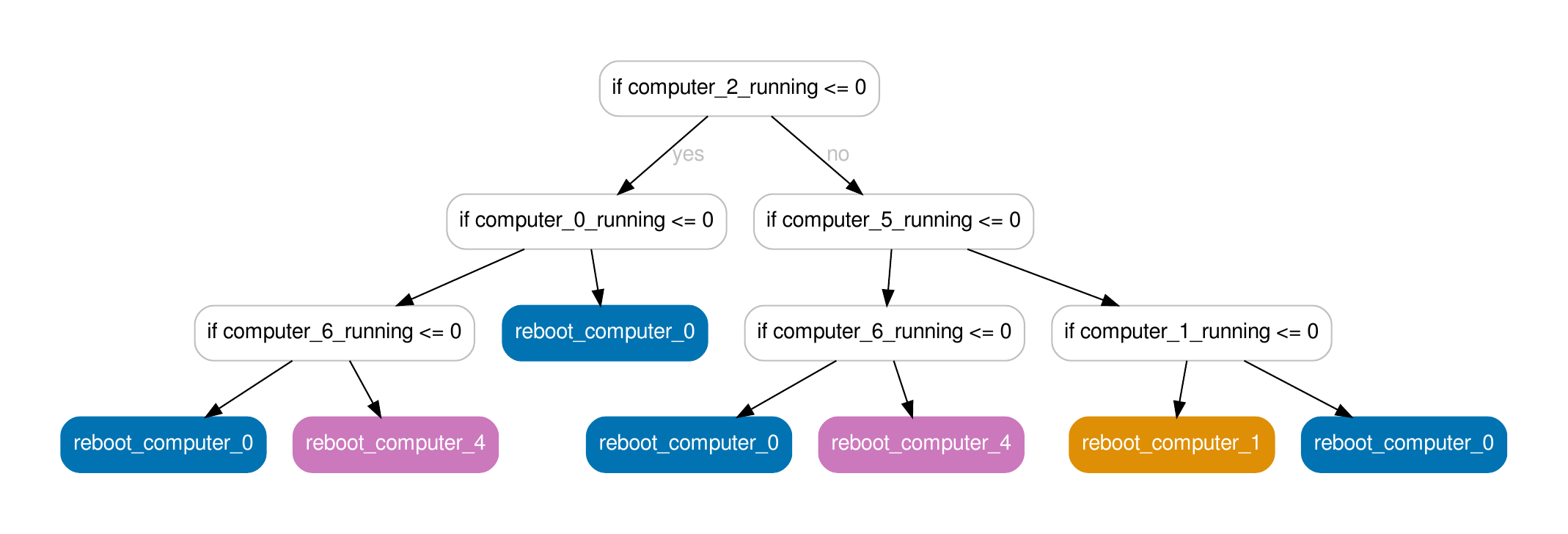}
        \caption{system\_administrator\_1}
    \end{subfigure}
    \hfill
    \begin{subfigure}[b]{0.32\linewidth}
        \centering
        \includegraphics[width=\textwidth]{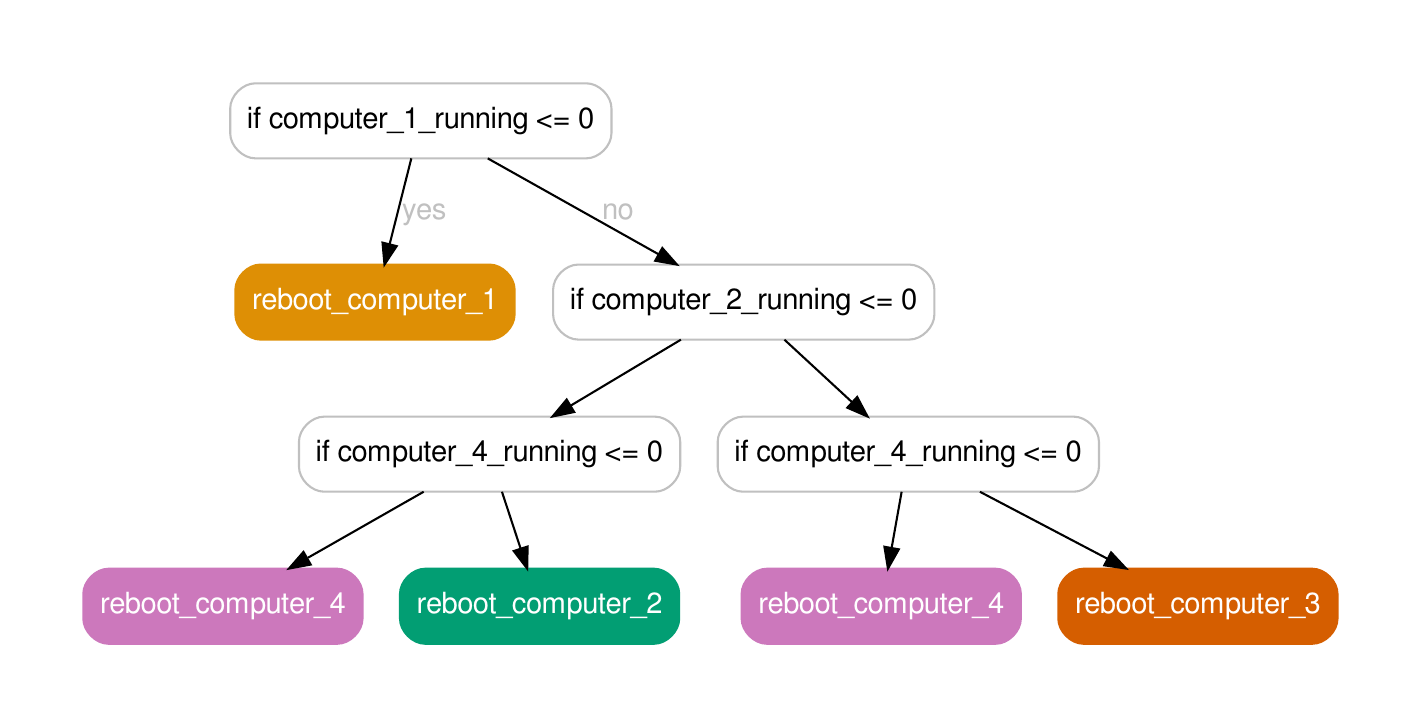}
        \caption{system\_administrator\_2}
    \end{subfigure}
    \hfill
    \begin{subfigure}[b]{.32\linewidth}
        \centering
        \includegraphics[width=\textwidth]{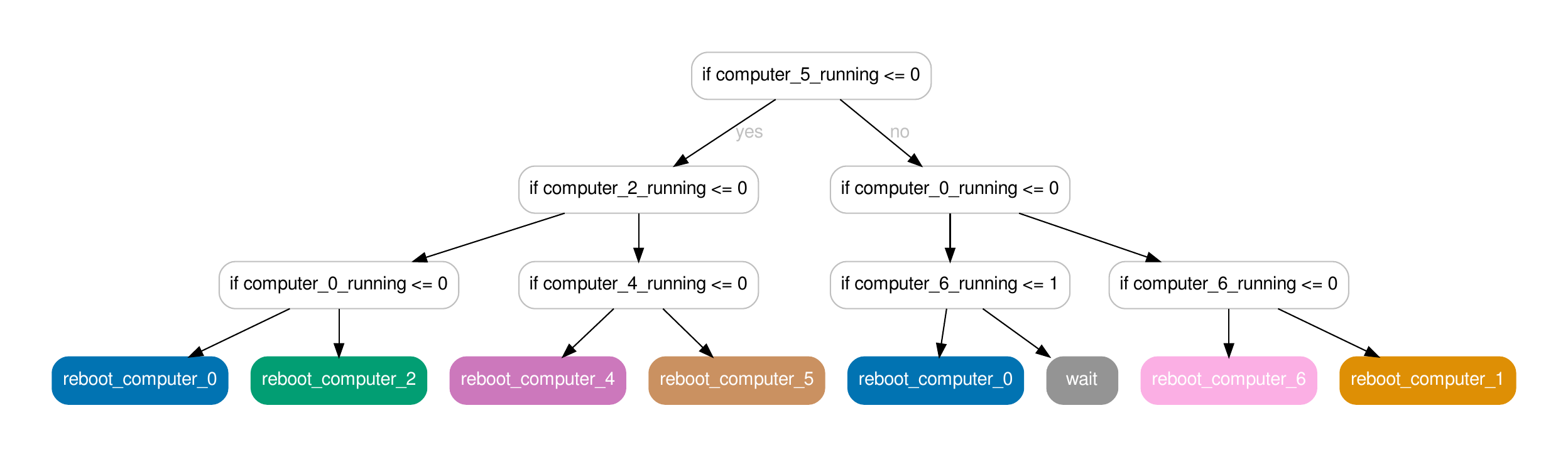}
        \caption{system\_administrator\_tree}
    \end{subfigure}
    \newline
    \begin{subfigure}[b]{.32\linewidth}
        \centering
        \includegraphics[width=\textwidth]{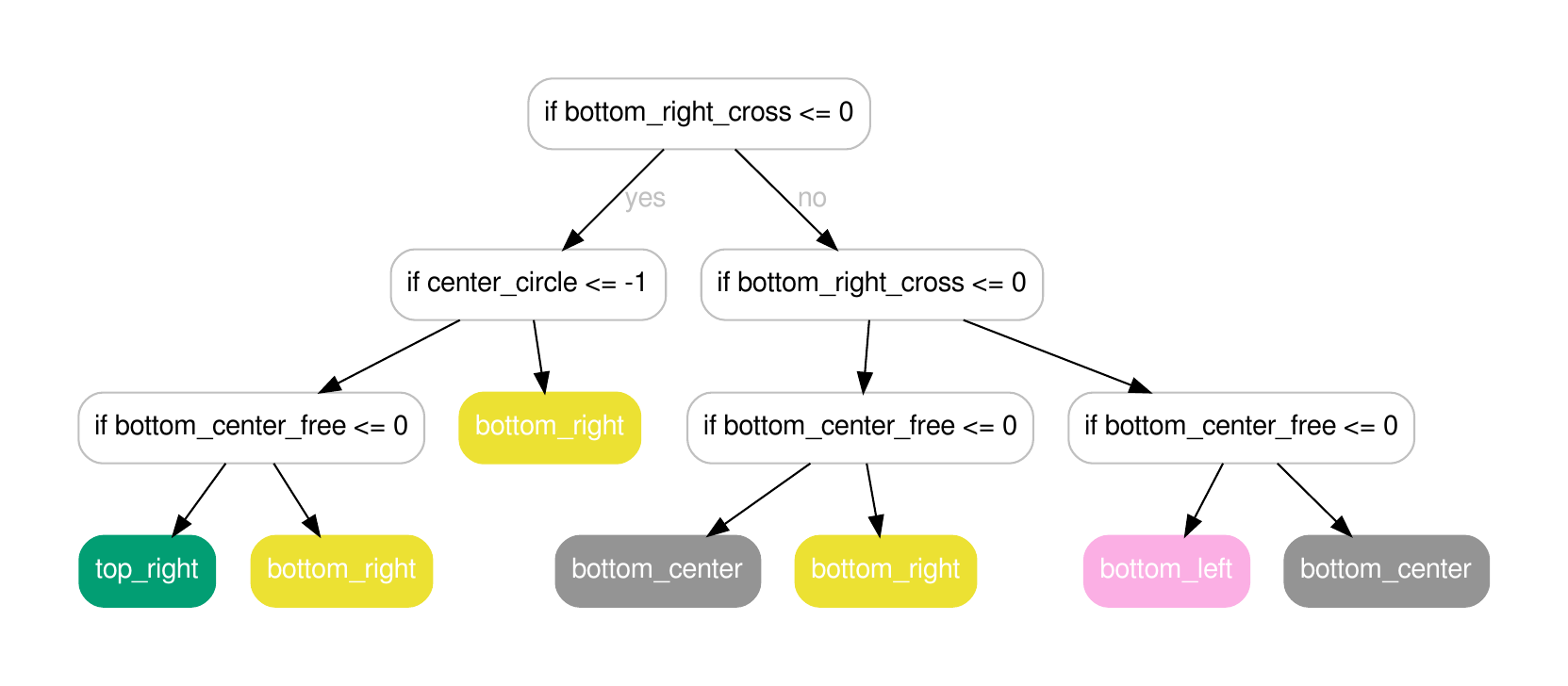}
        \caption{tictactoe\_vs\_random}
    \end{subfigure}
    \hfill
    \begin{subfigure}[b]{0.32\linewidth}
        \centering
        \includegraphics[width=\textwidth]{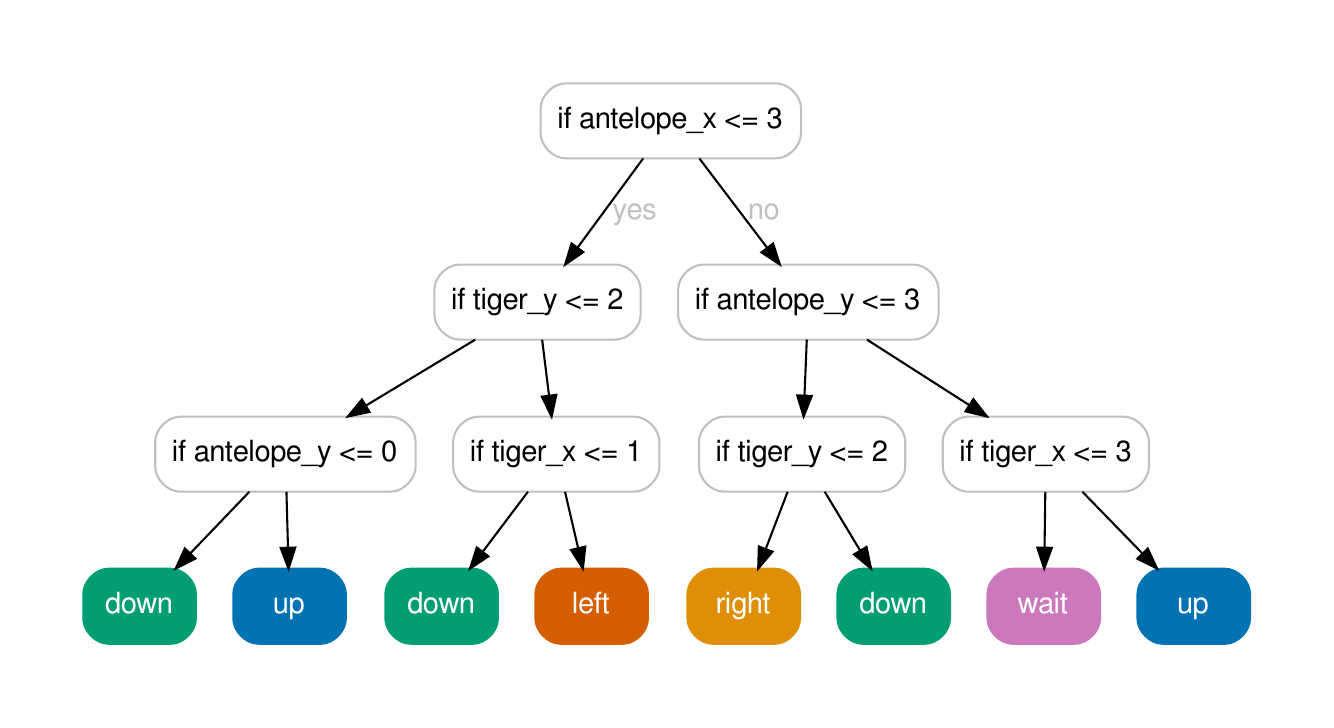}
        \caption{tiger\_vs\_antelope}
    \end{subfigure}
    \hfill
    \begin{subfigure}[b]{.32\linewidth}
        \centering
        \includegraphics[width=\textwidth]{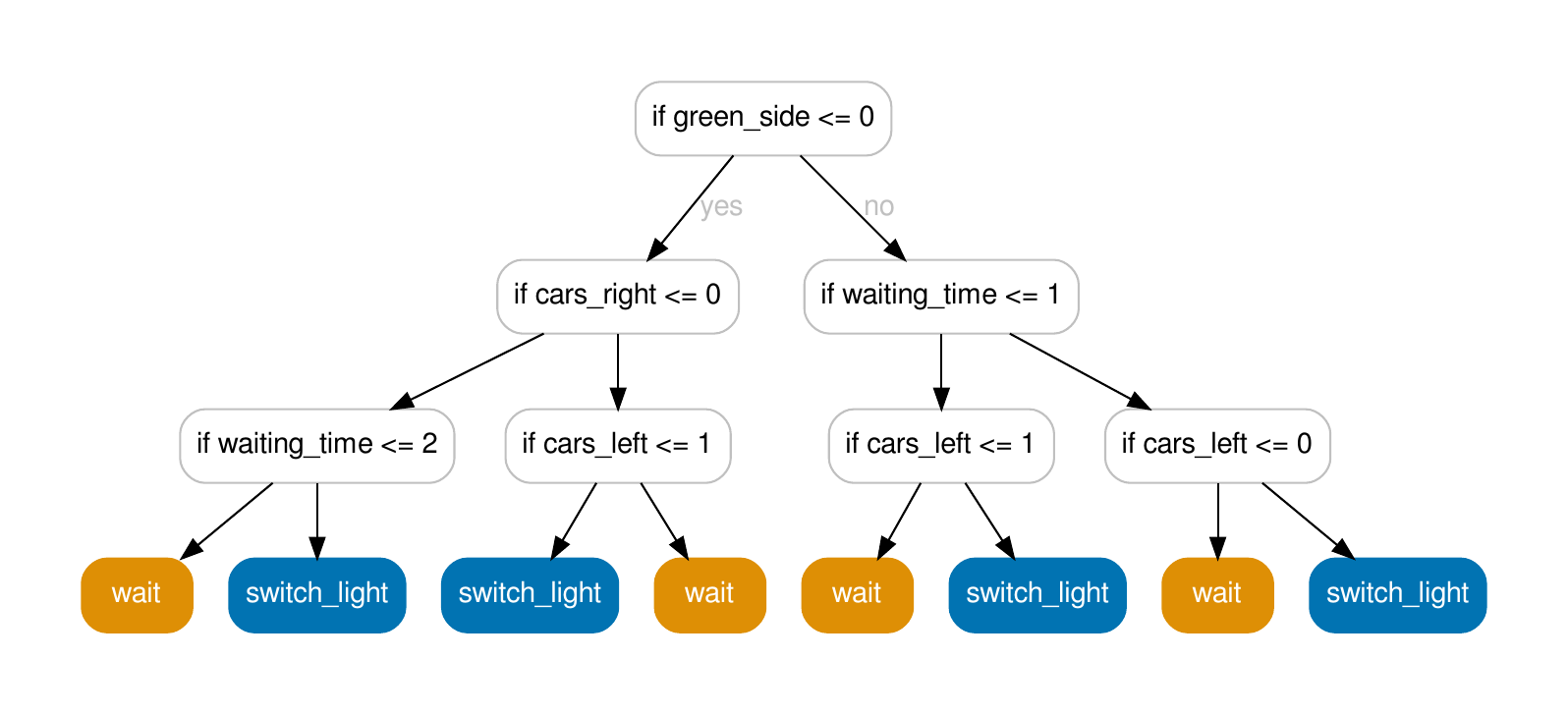}
        \caption{traffic\_intersection}
    \end{subfigure}
    \newline
    \begin{subfigure}[b]{.32\linewidth}
        \centering
        \includegraphics[width=\textwidth]{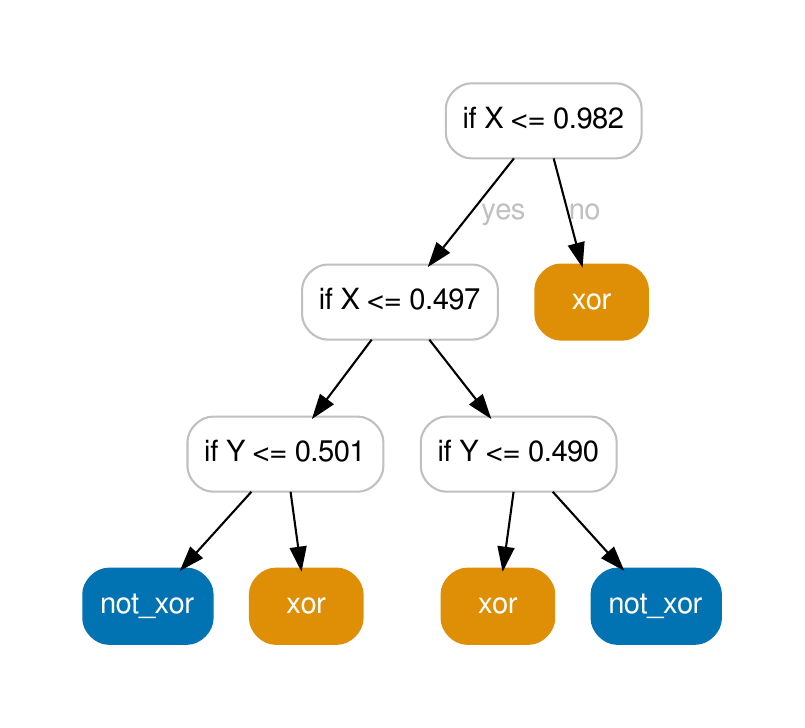}
        \caption{xor}
    \end{subfigure}
    \caption{
    Depth 3 trees produced by OMDT within 2 hours of runtime with one fixed seed.
    }
    \label{fig:tree-policies-omdt}
\end{figure*}

\begin{figure*}[p]
    \centering
    \begin{subfigure}[b]{.31\linewidth}
        \centering
        \includegraphics[width=\textwidth]{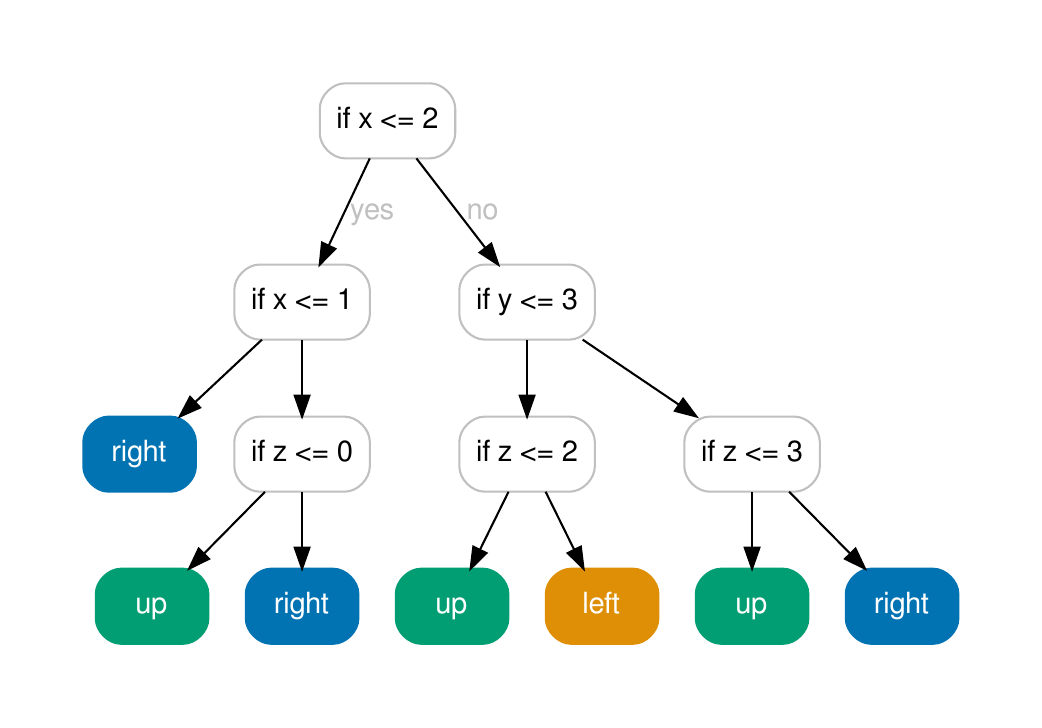}
        \caption{3d\_navigation}
    \end{subfigure}
    \hfill
    \begin{subfigure}[b]{0.31\linewidth}
        \centering
        \includegraphics[width=\textwidth]{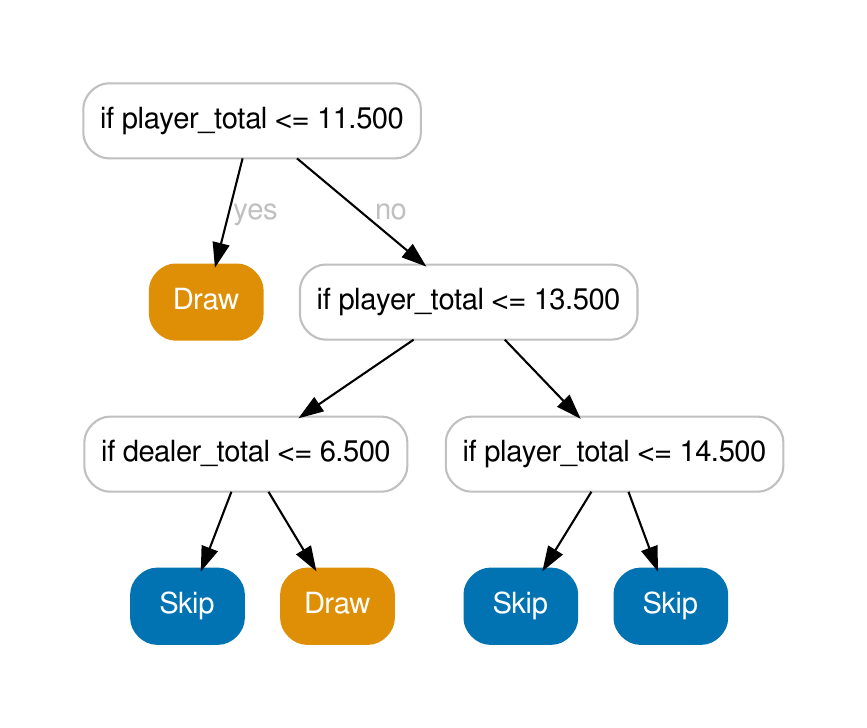}
        \caption{blackjack}
    \end{subfigure}
    \hfill
    \begin{subfigure}[b]{.31\linewidth}
        \centering
        \includegraphics[width=\textwidth]{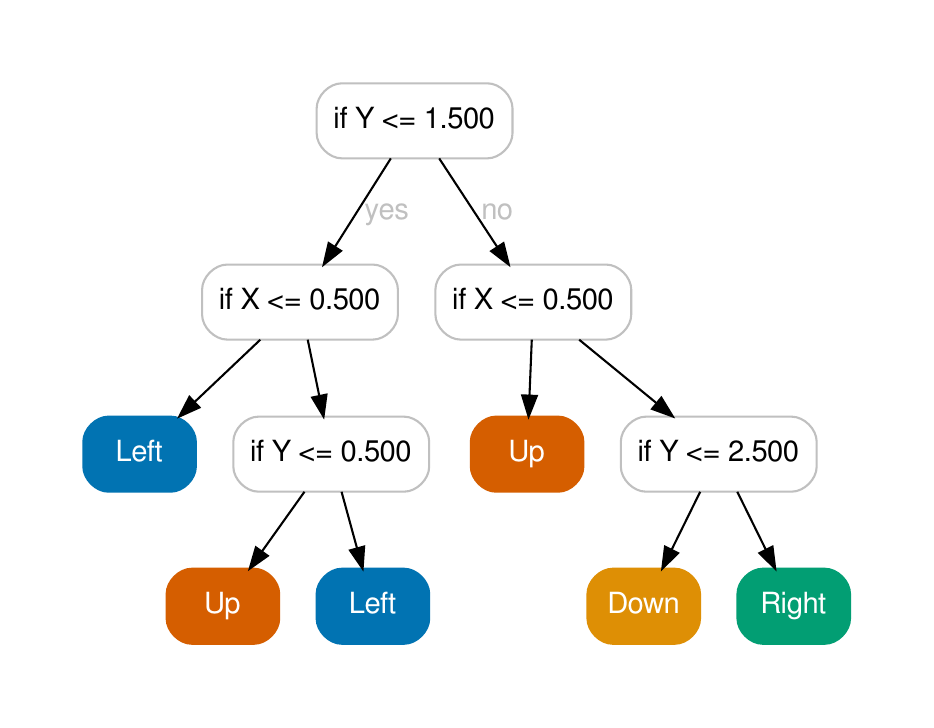}
        \caption{frozenlake\_4x4}
    \end{subfigure}
    \newline
    \begin{subfigure}[b]{.31\linewidth}
        \centering
        \includegraphics[width=\textwidth]{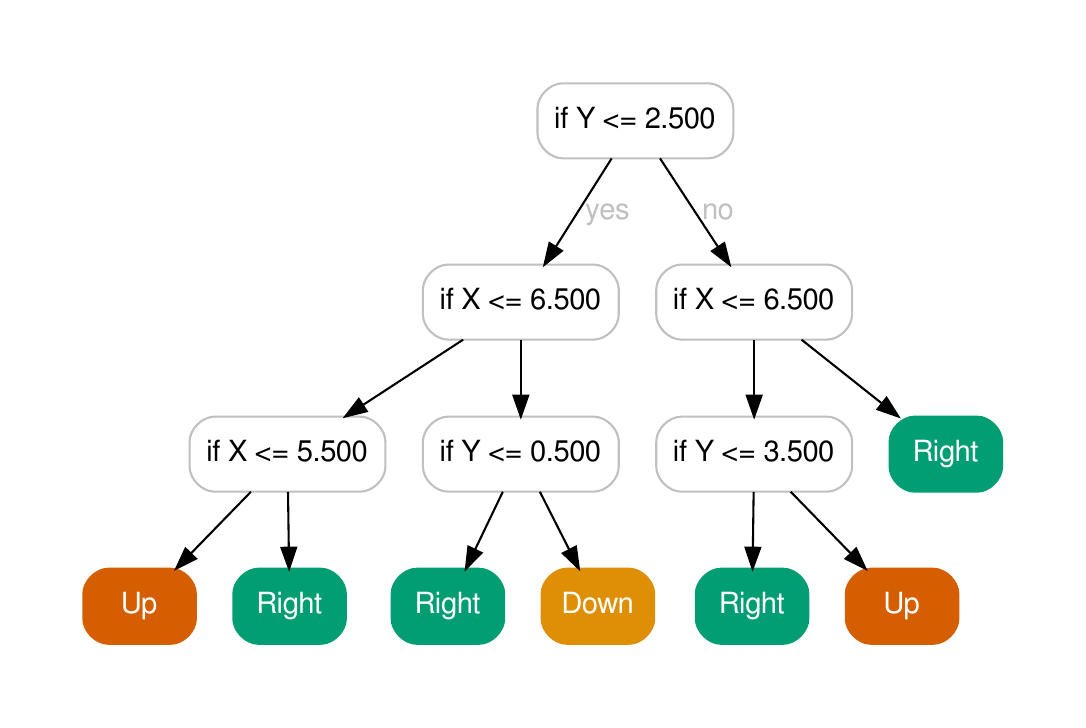}
        \caption{frozenlake\_8x8}
    \end{subfigure}
    \hfill
    \begin{subfigure}[b]{0.31\linewidth}
        \centering
        \includegraphics[width=\textwidth]{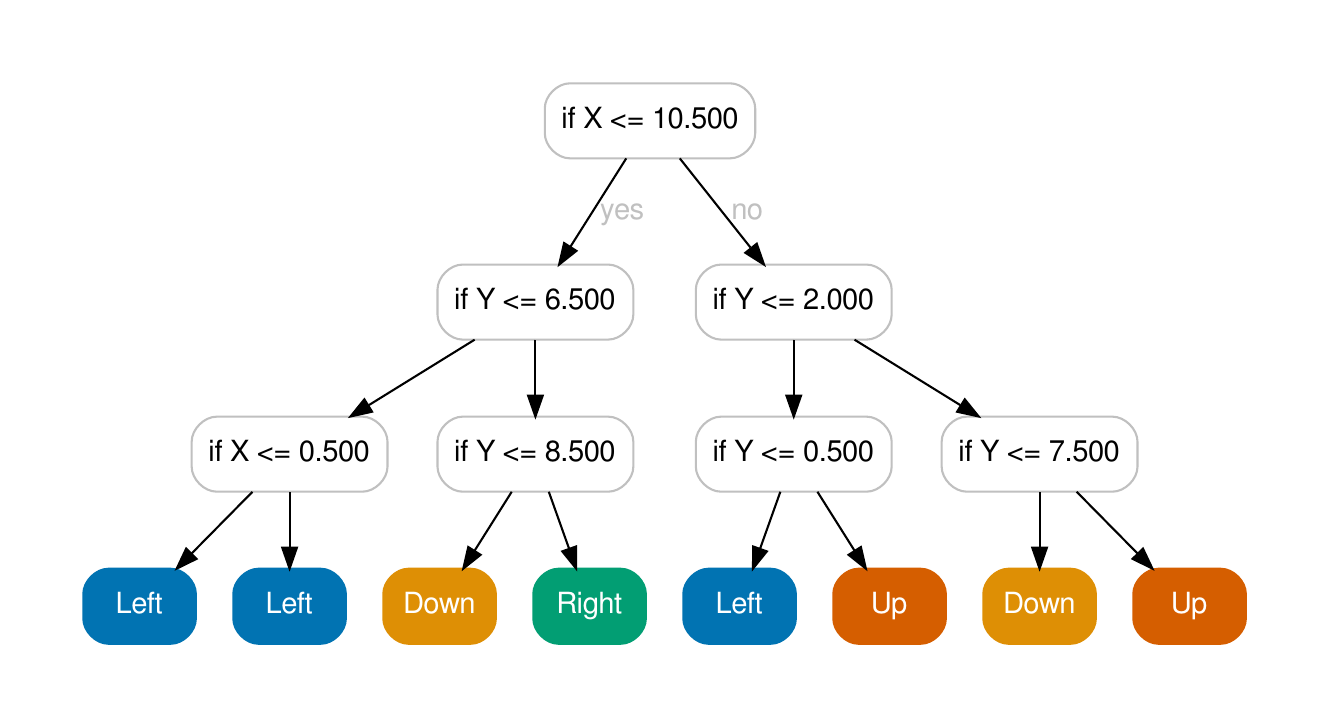}
        \caption{frozenlake\_12x12}
    \end{subfigure}
    \hfill
    \begin{subfigure}[b]{.31\linewidth}
        \centering
        \includegraphics[width=\textwidth]{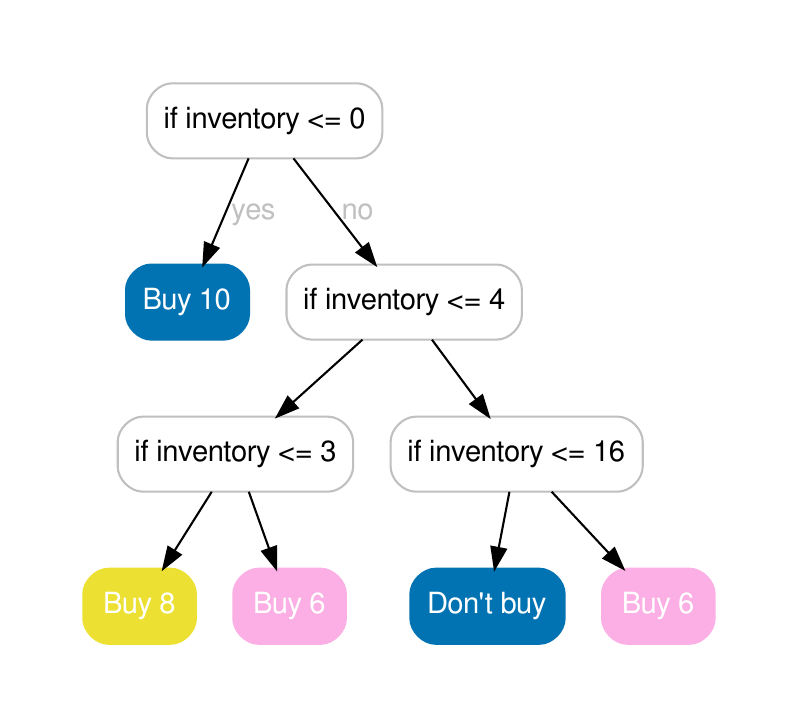}
        \caption{inventory\_management}
    \end{subfigure}
    \newline
    \begin{subfigure}[b]{.32\linewidth}
        \centering
        \includegraphics[width=\textwidth]{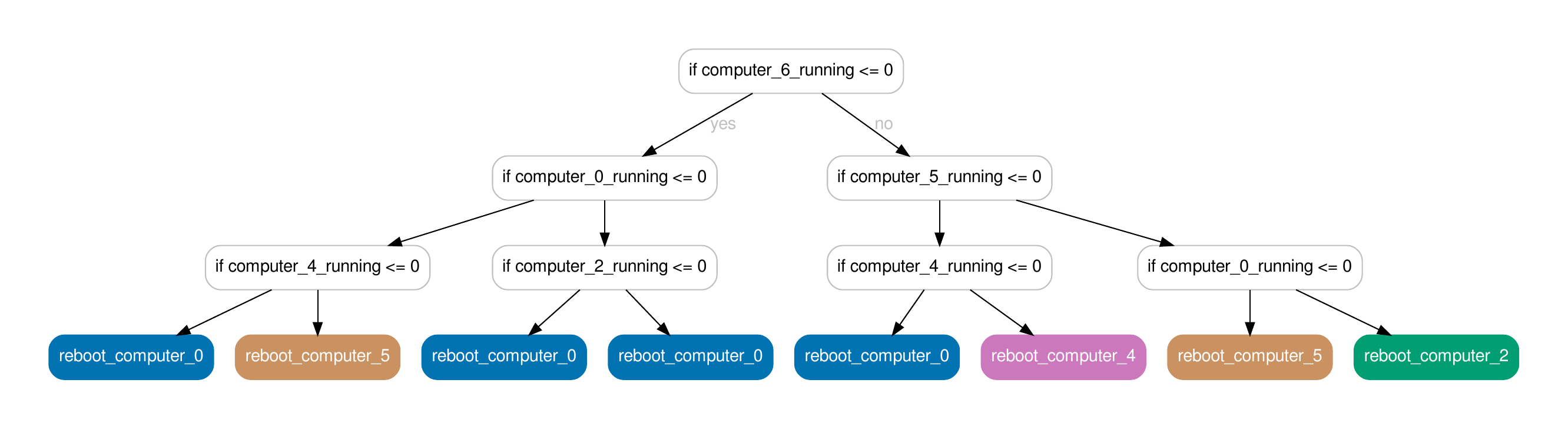}
        \caption{system\_administrator\_1}
    \end{subfigure}
    \hfill
    \begin{subfigure}[b]{0.32\linewidth}
        \centering
        \includegraphics[width=\textwidth]{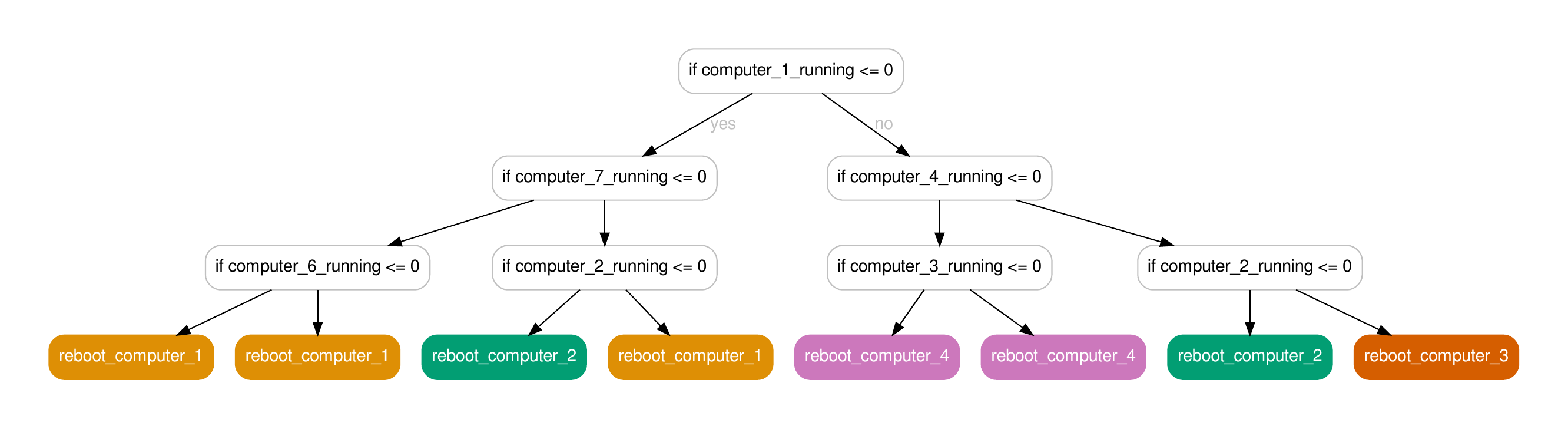}
        \caption{system\_administrator\_2}
    \end{subfigure}
    \hfill
    \begin{subfigure}[b]{.32\linewidth}
        \centering
        \includegraphics[width=\textwidth]{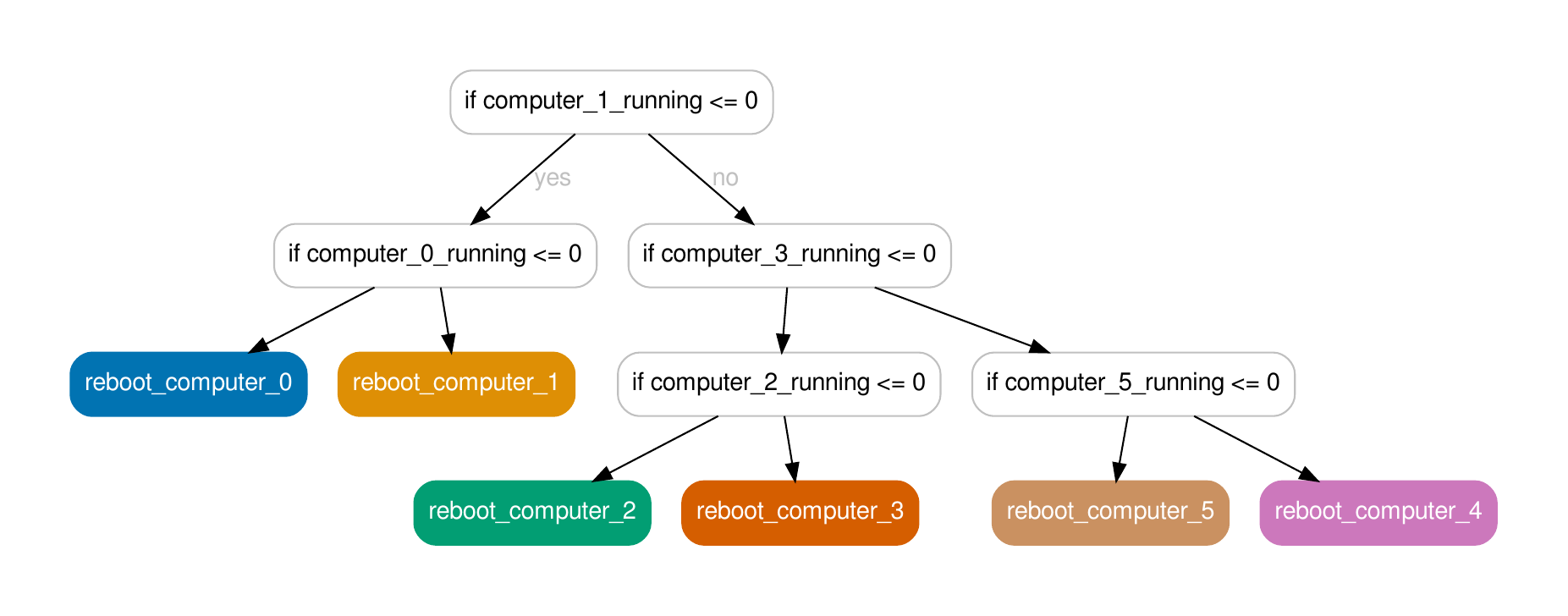}
        \caption{system\_administrator\_tree}
    \end{subfigure}
    \newline
    \begin{subfigure}[b]{.31\linewidth}
        \centering
        \includegraphics[width=\textwidth]{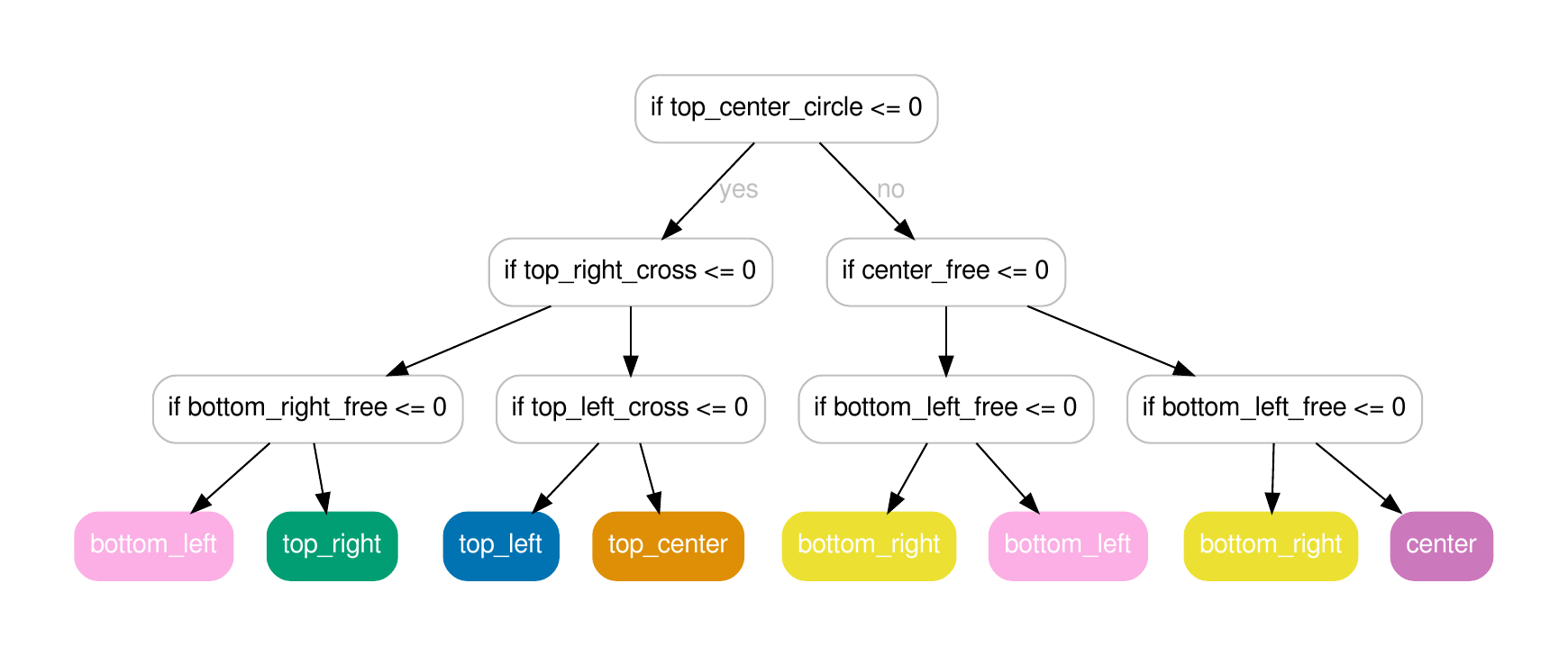}
        \caption{tictactoe\_vs\_random}
    \end{subfigure}
    \hfill
    \begin{subfigure}[b]{0.31\linewidth}
        \centering
        \includegraphics[width=\textwidth]{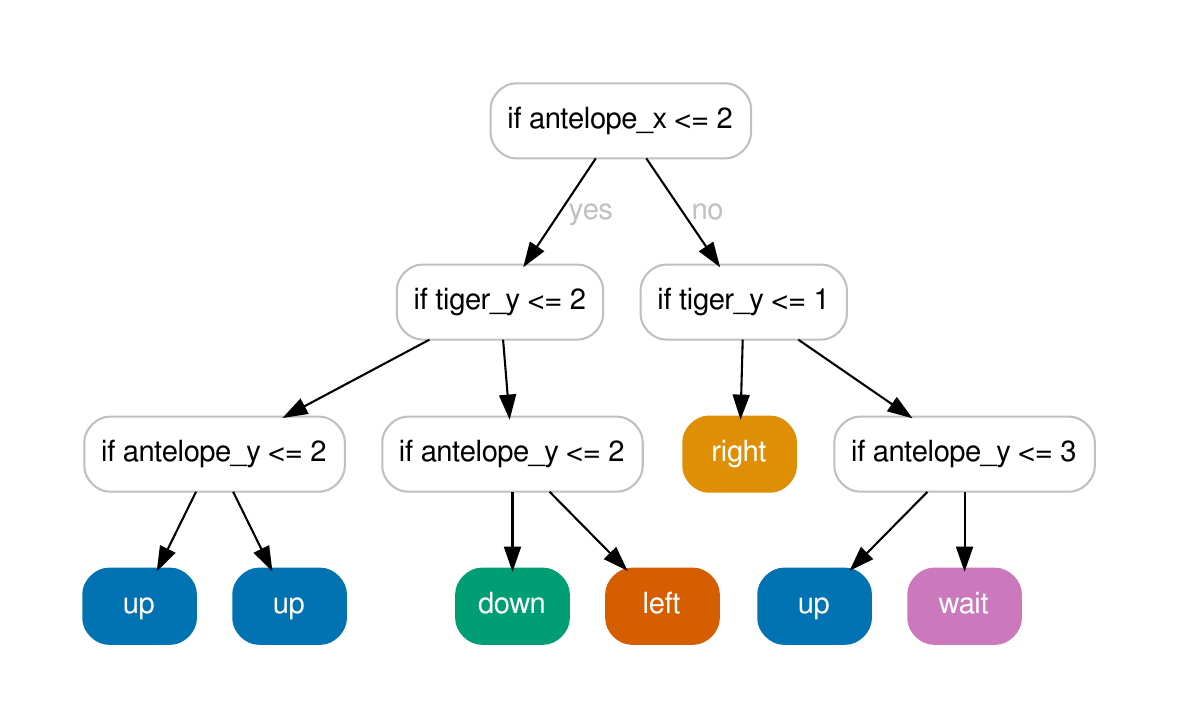}
        \caption{tiger\_vs\_antelope}
    \end{subfigure}
    \hfill
    \begin{subfigure}[b]{.31\linewidth}
        \centering
        \includegraphics[width=\textwidth]{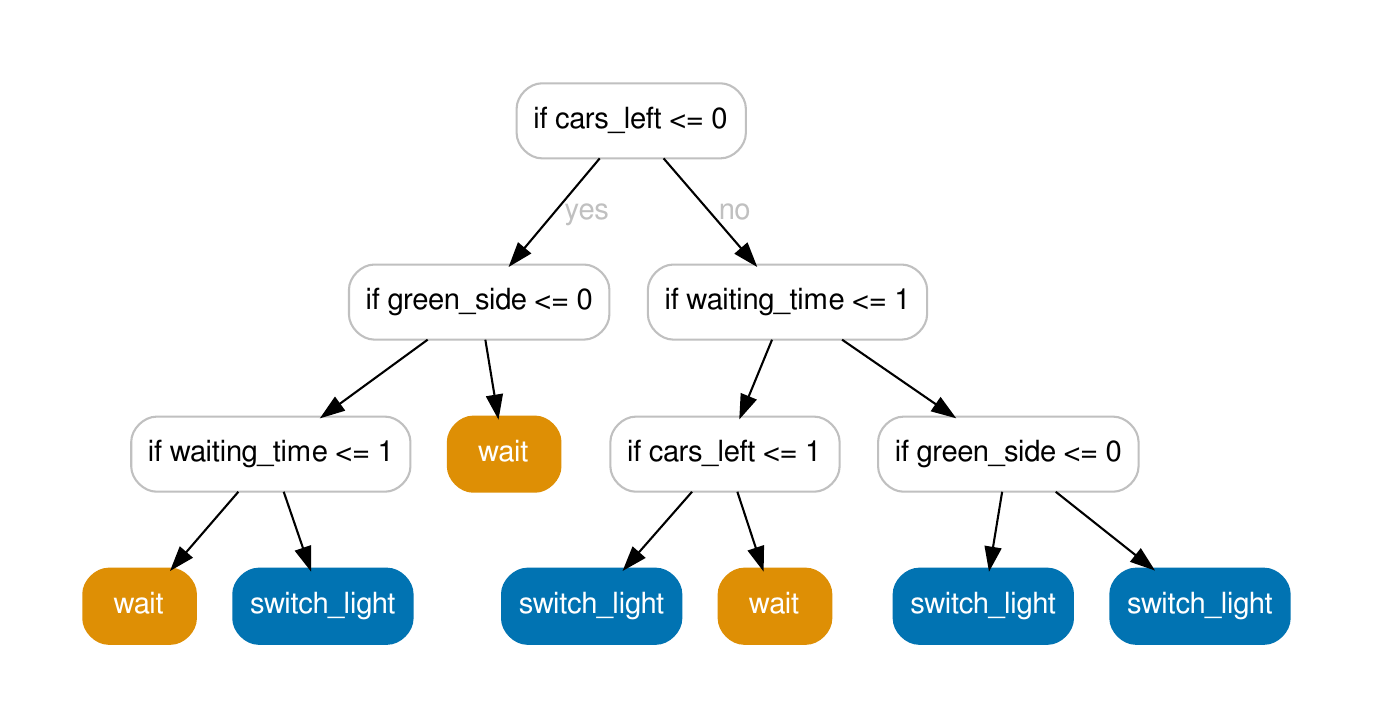}
        \caption{traffic\_intersection}
    \end{subfigure}
    \newline
    \begin{subfigure}[b]{.31\linewidth}
        \centering
        \includegraphics[width=\textwidth]{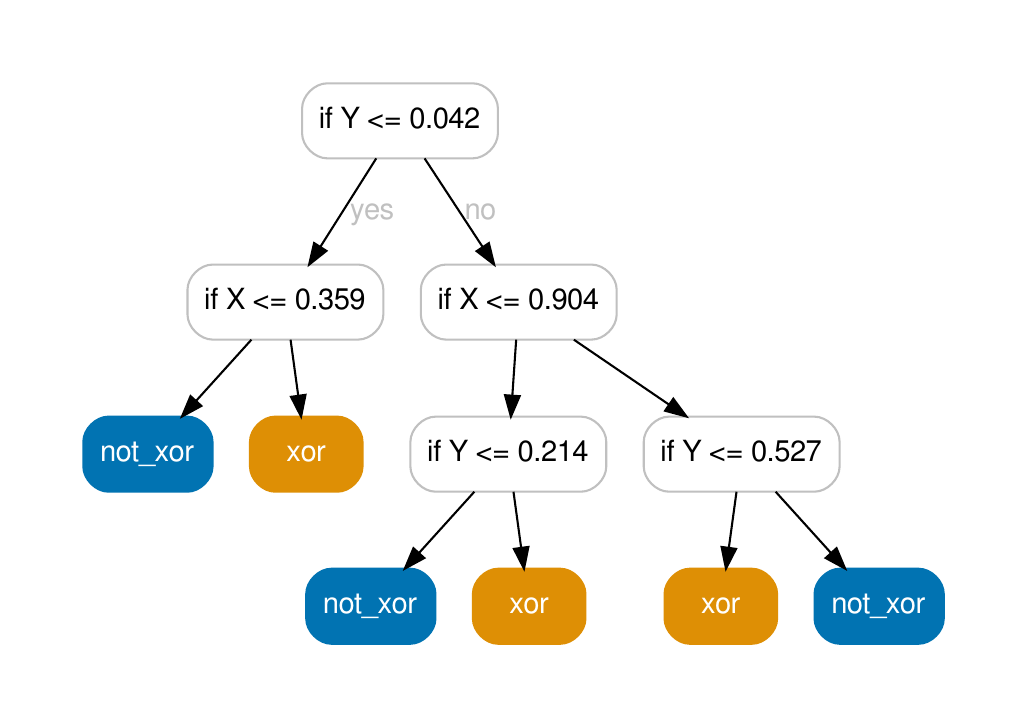}
        \caption{xor}
    \end{subfigure}
    \caption{
    Depth 3 trees produced by VIPER with one fixed seed.
    }
    \label{fig:tree-policies-viper}
\end{figure*}

We claim that since our models are interpretable since they are limited in size to an extent that a human can easily follow the predictions of the models~\cite{lipton2018mythos}. The decision trees of depth 3 created by OMDT and VIPER are visualized for inspection in Figure~\ref{fig:tree-policies-omdt} and Figure~\ref{fig:tree-policies-viper} respectively.

\end{document}